\documentclass[conference]{IEEEtran}
\IEEEoverridecommandlockouts
\usepackage{cite}
\usepackage{amsmath,amssymb,amsfonts}
\usepackage{algorithmic}
\usepackage{graphicx}
\usepackage{textcomp}
\usepackage{xcolor}
\usepackage{booktabs}
\usepackage{array}
\usepackage{placeins}
\usepackage{url}
\usepackage{longtable} 

\def\BibTeX{{\rm B\kern-.05em{\sc i\kern-.025em b}\kern-.08em
    T\kern-.1667em\lower.7ex\hbox{E}\kern-.125emX}}
\begin{document}


\title{Chasing Ghosts: A Simulation-to-Real Olfactory Navigation Stack with Optional Vision Augmentation}

%

\author{\IEEEauthorblockN{Kordel K. France}
\IEEEauthorblockA{\textit{Dept. of Computer Science} \\
\textit{University of Texas at Dallas}\\
Richardson, TX, USA \\
kordel.france@utdallas.edu}
\and
\IEEEauthorblockN{Ovidiu Daescu}
\IEEEauthorblockA{\textit{Dept. of Computer Science} \\
\textit{University of Texas at Dallas}\\
Richardson, TX, USA \\
ovidiu.daescu@utdallas.edu}
\and
\IEEEauthorblockN{Latifur Khan}
\IEEEauthorblockA{\textit{Dept. of Computer Science} \\
\textit{University of Texas at Dallas}\\
Richardson, TX, USA \\
latifur.khan@utdallas.edu}
\and
\IEEEauthorblockN{Rohith Peddi}
\IEEEauthorblockA{\textit{Dept. of Computer Science} \\
\textit{University of Texas at Dallas}\\
Richardson, TX, USA \\
rohith.peddi@utdallas.edu}
}

\pagestyle{plain}

\maketitle

\begin{abstract}
Autonomous odor source localization remains a challenging problem for aerial robots due to turbulent airflow, sparse and delayed sensory signals, and strict payload and compute constraints.
While prior unmanned aerial vehicle (UAV)-based olfaction systems have demonstrated gas distribution mapping or reactive plume tracing, they rely on predefined coverage patterns, external infrastructure, or extensive sensing and coordination.
In this work, we present a complete, open-source UAV system for online odor source localization using a minimal sensor suite. 
The system integrates custom olfaction hardware, onboard sensing, and a learning-based navigation policy that we train in simulation and deploy on a real quadrotor. 
Through our minimal framework, the UAV is able to navigate directly toward an odor source without constructing an explicit gas distribution map or relying on external positioning systems. 
We incorporate vision as an optional complementary modality to accelerate navigation under certain conditions.
We validate the proposed system through real-world flight experiments in a large indoor environment using an ethanol source, demonstrating consistent source-finding behavior under realistic airflow conditions. 
The primary contribution of this work is a reproducible system and methodological framework for UAV-based olfactory navigation and source finding under minimal sensing assumptions. 
We elaborate on our hardware design and open source our UAV firmware, simulation code, olfaction-vision dataset, and circuit board to the community.\footnote{Code, data, and designs available at \url{https://github.com/KordelFranceTech/ChasingGhosts}}
\end{abstract}

\maketitle

\section{Introduction} 
How does one find what it cannot see, hear or touch?
Researchers have likened navigating by scent to "chasing ghosts" due to humans' lack of high bandwidth olfactory perception.
Olfaction is the most primitive form of perception, yet artificial intelligence systems have predominantly focused on visual, audio, and linguistic data.
This oversight largely stems from the scarcity of olfactory data and the absence of standardized benchmarks, which pose significant challenges for developing and evaluating machine learning models in this domain.

\begin{figure}
  \centering
  \includegraphics[width=85mm]{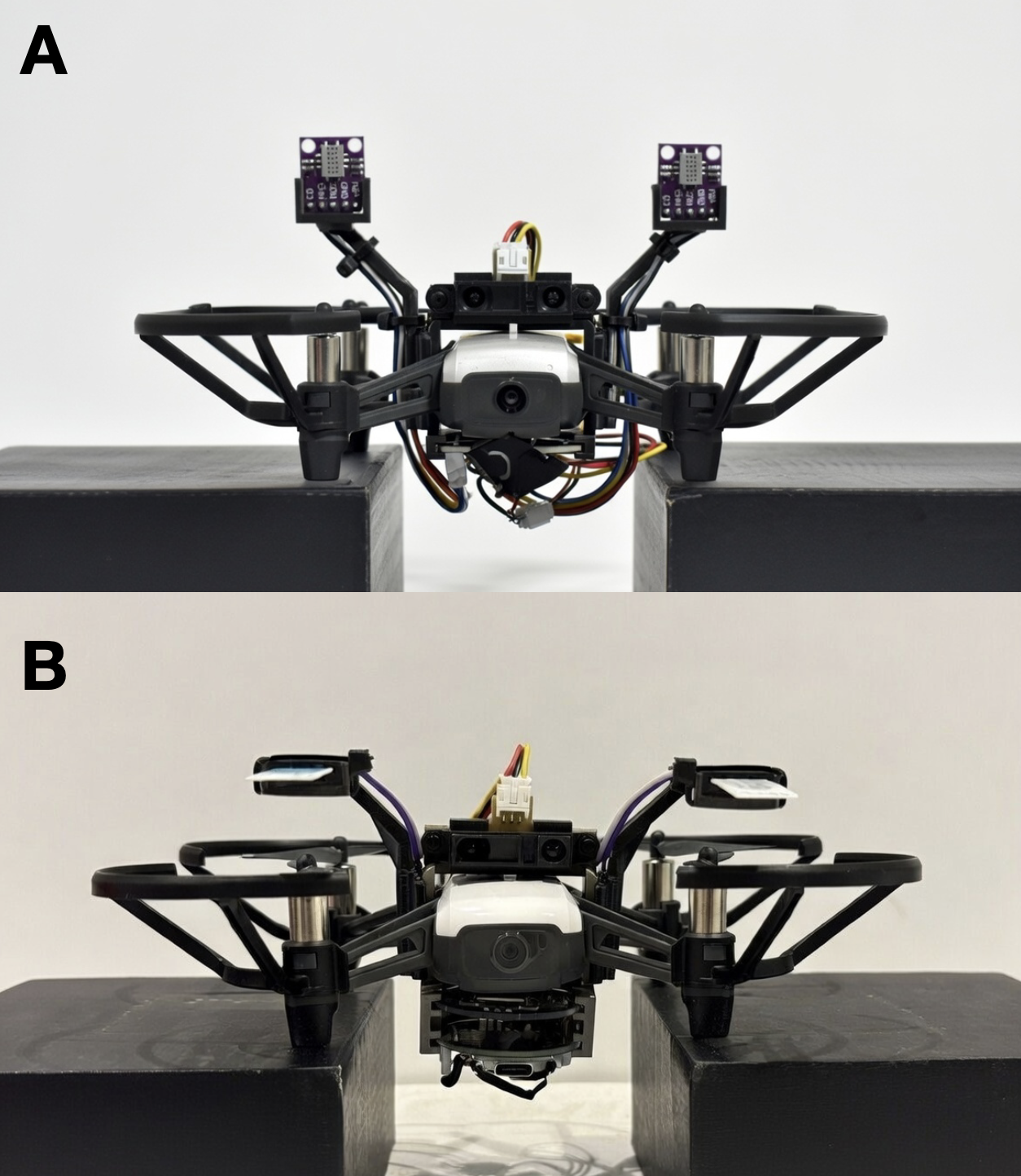}
  \caption{The UAV equipped with the olfactory processing unit (OPU) and sensor harnesses for scent-based navigation. Panel A shows the metal-oxide sensor configuration, and panel B shows the electrochemical sensor configuration.}
  \label{fig:uavmain}
\end{figure}

In this study, we introduce a model for olfaction-only navigation, but demonstrate how vision can complement performance.
We show how these models can be trained via simulation to run at the edge for scent-based navigation to an odor source.
Olfactory navigation is largely absent from robotic functions today, and our motivation for this work is to establish methodology that allows robots to, for example, localize the source of a specified chemical compound, inform automated surgery by mapping olfactory cues to visual components, and inspire alternative forms of underwater navigation via chemical plume tracking.
We demonstrate our methodology on a real UAV navigating to a chemical compound.

Due to the hysteresis and non-linearity associated with gas dynamics, navigating by scent is inherently difficult. 
Current olfaction sensors suffer from drift and long sampling times \cite{dennler22drift, dennler_limitations_2024, france2025_fast_chronamp}, and the inability to accurately perceive the presence of a chemical compound analogizes to "chasing ghosts".
Here, we detail a confluence of adaptive learning techniques, hardware design, sensor selection, and compute optimizations that make this feasible.
We provide a detailed account of our hardware, methodologies, and evaluation metrics to ensure transparency and reproducibility. 
The Supplementary Material details a full bill of materials and code to reproduce our results.
Figure \ref{fig:uavmain} shows frontal views of the two configurations we used for the UAV.

The contributions of our work are as follows:
\begin{enumerate}
    \item A complete, open-source UAV system for odor source localization, integrating custom olfaction hardware, onboard sensing, simulation, and real-world deployment under payload and compute constraints.
    \item A learning-based navigation method that we train in simulation and deploy on a real UAV, enabling online odor source localization without prior mapping or external positioning infrastructure.
    \item A minimal-sensor navigation formulation, demonstrating reliable source-seeking behavior using only a small number of olfaction sensors and a single range sensor, with vision serving optionally as a complementary modality.
    \item Real-world experimental validation in a large indoor environment, showing consistent source localization behavior and providing qualitative and quantitative insights into plume navigation under realistic airflow conditions.
    \item A demonstration that our framework generalizes across sensing mediums.
\end{enumerate}

This work emphasizes system design and methodological integration, on which one could expand with statistically powered evaluation across a wider range of environments and conditions.

\section{Related Work}

Early UAV-based olfaction systems have focused on gas distribution mapping (GDM), where a robot follows a predefined flight pattern to reconstruct a concentration field and estimate a source location post-hoc.
Burgués et al. \cite{sniffybug-single-burgues19} present a representative example, employing a quadrotor to sample gas concentrations along a sweeping trajectory and infer source location from the reconstructed map. 
While effective for environmental monitoring, such approaches do not perform online source-seeking behavior and require substantial spatial coverage prior to localization.

Reactive plume tracing methods draw inspiration from biological odor tracking, enabling robots to move toward an odor source based on local concentration cues.
Shigaki et al. \cite{shigaki2018quadcopter} demonstrate three-dimensional chemical plume tracing using a quadcopter equipped with dual alcohol sensors and onboard ranging, achieving successful localization in controlled indoor environments. 
Their work highlights the feasibility of UAV-based plume tracking but focuses on reactive control strategies rather than learning-based policies trained across environments.


\begin{figure}
  \centering
  \includegraphics[width=85mm]{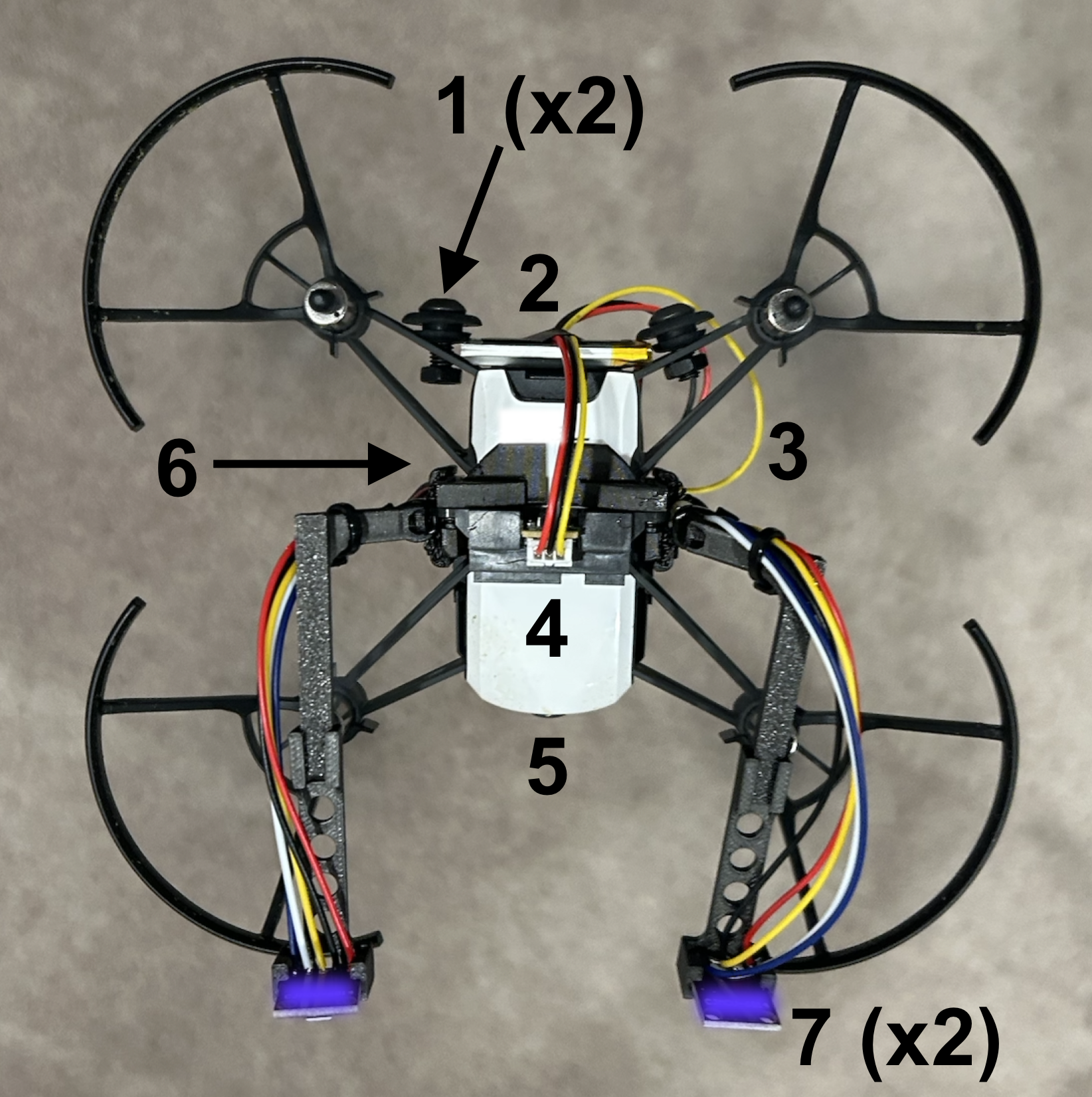}
  \caption{The UAV equipped with the olfactory processing unit (OPU) and sensor harnesses. The figure shows a top-down view of the aircraft: (1) the 2 ballast points needed to balance the aircraft; (2) the battery to power the motherboard required for the olfactory sensors; (3) the wire harnessing leading to the motherboard attached to the belly of the aircraft; (4) the forward-looking time-of-flight sensor for obstacle avoidance; (5) the forward-looking camera; (6) the motherboard and downward-looking time-of-flight infrared sensors attach to the belly (not shown); (7) the two olfactory sensor antennae (MOX sensors shown, but EC sensors lie at the same location when in the proper configuration).}
  \label{fig:uavviews}
\end{figure}

More recently, researchers have explored learning-based and swarm-based approaches to improve robustness in turbulent environments.
Duisterhof et al. \cite{sniffybug-swarm-duisterhof21} introduce a multi-agent “Sniffy Bug” system that combines bio-inspired navigation with particle swarm optimization, leveraging inter-agent communication and additional sensing modalities. 
While effective at the swarm level, such systems rely on greater sensing and coordination infrastructure than a single lightweight UAV typically carries.
Research from Zhang, et al. \cite{ZHANG2025_oevs, ZHANG2026_msaticnn} demonstrates how chemical and visual sensing can work collaboratively on a UAV, but does not demonstrate it in navigation tasks.

In contrast to prior work, this paper focuses on single-agent, learning-based odor source localization under minimal sensing assumptions. 
Rather than reconstructing gas distributions or relying on swarm coordination, the proposed system learns a navigation policy in simulation and deploys it directly on a real UAV, enabling online source-seeking behavior in a large indoor environment.
The emphasis is on system integration, reproducibility, and sim-to-real transfer, providing a foundation for future statistically powered evaluations and extensions.
The works in~\cite{sniffybug-single-burgues19, sniffybug-swarm-duisterhof21} leverage ultrawide bandwidth (UWB) radio tags placed throughout each course to help the UAV navigate. In contrast, we only allow the UAV to use olfaction (and at the last mile, vision) to guide itself to the odor source, and it has no prior knowledge about the course.
Unlike the post-mapping over $160m^2$ from \cite{sniffybug-single-burgues19}, we achieve real-time localization over $200m^2$, the largest size experimental course so far, without positional aids.

Work by Hassan, et al. \cite{hassan_localization_2024} shows how robots can navigate by olfactory and visual capabilities, but leverages closed-source online large language models to perform the computation as well as the interpretation of the chemical and vision data, methods impractical for real-time edge robotics.
In~\cite{sniffybug-single-burgues19, shigaki2018quadcopter, sniffybug-swarm-duisterhof21, hassan_localization_2024}, the authors use metal oxide sensors as the olfactory receptor.
We demonstrate our framework with metal oxide sensors to show improvement on previous work, but also demonstrate that our framework generalizes across sensing mediums with the introduction of electrochemical sensors.

Feng, et al. construct a dataset of vision-olfactory relationships in SmellNet \cite{feng2025smellnetlargescaledatasetrealworld}, but the dataset relates exclusively to fruits, nuts and other common foods.
Ozguroglu, et al. \cite{ozguroglu2025newyorksmellslarge} show how vision and olfaction can link together through contrastive learning, but the scope is exclusively to botany.
We build off many of the methods established by \cite{sniffybug-single-burgues19, sniffybug-swarm-duisterhof21, Singh2023}.
Namely, we use ethanol as our target compound to maintain parity with their work and provide a level of experimental control.

To the knowledge of the authors, this is the first demonstration of stereo olfactory sensing combined with vision on a UAV over two different chemical sensing mediums.
Section \ref{sec:sm_relatedwork} of the Supplementary Material presents a concise comparison of our work to related work.

\section{Methods}

Our primary goal is to demonstrate that UAVs can navigate purely with olfactory sensors to track a target chemical back to its source.
However, we recognize that visual sensors can complement olfactory navigation as seen in several mammals and insects.
As a result, we show how understanding the aromas emitted by certain objects can decrease the time needed to localize the target compound through visual confirmation of capturing the odor source.

\subsection{Hardware}

We desire a solution for olfactory navigation that is financially and technically accessible to robotics and AI researchers while also reaching the precision required to localize a chemical compound.
Therefore, we select off-the-shelf hardware where reasonably possible.
Figure \ref{fig:uavviews} shows critical additions to the base UAV airframe, while Section \ref{sec:sm_uavcad} of the Supplementary Material discusses a full bill of materials.

We selected the DJI Tello as the UAV \cite{DJITelloPy}, a small open-sourced drone that one can program, but which contains minimal onboard compute.
Unlike prior works \cite{sniffybug-single-burgues19, sniffybug-swarm-duisterhof21} that leverage the Crazyflie UAV platform, we require a slightly larger UAV platform with a camera and higher payload that enables us to carry more sensing capability and compute.
Three infrared sensors provide basic stability: two downward-looking sensors for leveling and one forward-looking sensor for obstacle detection.
We designed a custom olfactory processing unit (OPU) that contains an ESP-32 microcontroller and a PalmSens EmStat Pico potentiostat and attaches to the bottom of the UAV.
The OPU processes the infrared data for obstacle detection and olfaction sensors to inform navigation, and wirelessly streams all data back to a laptop ground station for telemetry.
We mechanically modify the Tello UAV to accommodate the harnessing, battery, and circuitry the OPU requires.
This required adding 71 grams of ballast to the aft end of the drone in order to counter the forward moment the additional components create; no additional ballast was needed to adjust for lateral moments.
We designed the mechanical brackets and harnessing needed to attach the OPU, olfactory sensors, and infrared sensors to the airframe in SolidWorks \cite{solidworks2024}.
We designed the OPU motherboards using EasyEda \cite{easyeda2025}.
Sections \ref{sec:sm_uavbrakout} and \ref{sec:sm_uavcad} present details on the electrical and mechanical hardware, respectively.

\begin{figure}
  \centering
  \includegraphics[width=88mm]{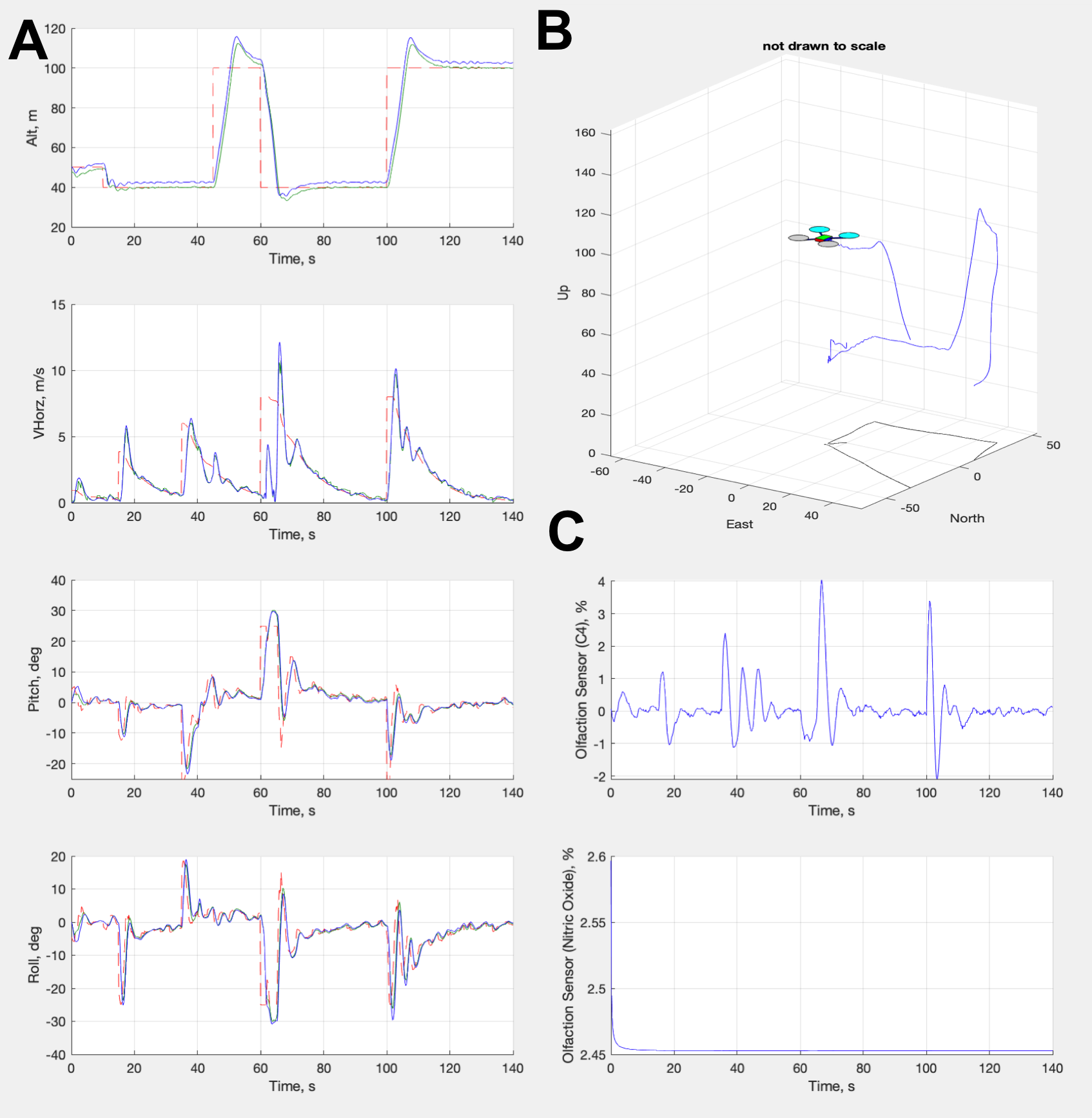}
  \caption{(A) From top to bottom, this panel shows the altitude, velocity, pitch, and roll from the controller response (blue line) and the command (red line) from the controls algorithms. (B) A diagram showing a partial flight path of the UAV in simulation. (C) From top to bottom, this panel shows the olfactory signal responses for ethanol and nitrogen dioxide.}
  \label{fig:control_sim}
\end{figure}

\subsection{Simulation}
We created a digital twin of the modified UAV for two different simulations. We built a physics simulation via SimuLink to model the control loop algorithms for the UAV, test various flight conditions, and establish a control envelope.
Figure \ref{fig:control_sim} shows a series of plots from this simulation, establishing the performance of the autopilot control algorithms (for more information on these algorithms, see Section \ref{sec:sm_autopilot} and Tables \ref{tab:autopilot_table} and \ref{tab:autopilot_gains} in the Supplementary Material).
This simulation contains a configuration file that enables extrapolation of our control laws to any other UAV platform with our OPU modification.
We built the second simulation using Python and the \textit{Gymnasium} framework \cite{Brockman2016:DBLP:journals/corr/BrockmanCPSSTZ16}, which we discuss in detail in the next sections.
We leverage and tune Kalman filters to smooth signals in control responses.
Taking inspiration from olfactory navigation in insects, we position two sensors at the top and fore of the UAV in analogous positions as antennae on the silk moth. 
This differs from previous approaches of olfactory navigation \cite{sniffybug-single-burgues19, sniffybug-swarm-duisterhof21} where the sensor sat at the top of the body and below the rotors, but is similar to the locations proposed by \cite{shigaki2018quadcopter}.
Our positioning allows us to perform stereo sensing to ascertain plume direction and avoid volatility from the rotor wash, maximizing the odor signal.

\subsection{Sensors}
We desire a generalized solution for olfactory navigation.
To aid with down-selection of off-the-shelf olfaction sensors, we designed a small electrical breakout board (see Section \ref{sec:sm_uavbrakout} in the Supplementary Material).
This allowed us to manually test detection algorithms without having to re-configure the mounting hardware and ballast on the drone.
The repository associated with this paper contains the circuitry and design of this board.
The results of the experiments we performed with this olfactory board showed that metal oxide (MOX) and electrochemical (EC) sensors were the best for tracking ethanol, so we selected them as the final olfactory effectors.

Metal oxide sensors respond rapidly (1-100 Hz) \cite{dennler_high-speed_2024} and are typically sensitive to \text{families} of molecules, allowing for more general chemical tracking.
We select a pair of Sensirion MICS 6814 sensors due to their prevalence in recent literature \cite{dennler_high-speed_2024, feng2025smellnetlargescaledatasetrealworld} and the wide array of compounds they detect.
These sensors respond to a variety of gases, but we focus on their sensitivity to ethanol for our work here since ethanol is the target compound.

Metal oxide sensors change resistance based on their exposure to the target gas.
This resistance indicates gas intensity in the surrounding environment and one can compute it according to the following equation:

\begin{equation}
    R_s = \left [ \frac{V_c}{V_{RL}} - 1\right ] \cdot R_L
    \label{eq:mox}
\end{equation}

\noindent 
where $R_S$ indicates the sensor resistance, $V_c$ is the circuit voltage, $V_{RL}$ is the voltage drop over the load resistor, and $R_L$ is the resistance of said load resistor.
As Equation \ref{eq:mox} is fundamental to the function of the MOX sensor, we design our optimizations around it.
We build off of the methodology from Burgués, et al. in \cite{sniffybug-single-burgues19} in that we design our algorithm to not depend on absolute concentrations in part-per-million or part-per-billion quantity.
Rather, we analyze the relative change between time steps of $V_{RL}$ (the temporal difference) and use this to inform the UAV on how to move relative to gas measurements.

Electrochemical sensors respond less rapidly, but much more specifically to individual compounds, allowing for more precise tracking.
We leverage a two-electrode electrochemical sensor from ItalSens designed to maximize diffusivity of compounds over a certain voltage range for chronoamperometry.
Chronoamperometry is the process by which one can measure the change in electric current with respect to time over a controlled electrical potential.
The Cottrell equation shows this relationship: 

\begin{equation}
    I = \frac{n_eFAc_k \sqrt{D_k}}{\sqrt{\pi t}}
    \label{eq:cottrell}
\end{equation}

\noindent where $I$ denotes the electrical current, measured in amperes; $n_e$ is the number of electrons needed to oxidize one molecule of analyte $k$; $F$ is the Faraday constant of 96,485 Coulombs per mol; $A$ denotes the planar area of the electrode in square centimeters; $c_k$ defines the initial concentration of the target analyte $k$ in mols per cubic centimeter; $D_k$ defines the diffusion coefficient for analyte $k$ in square centimeters per second; and $t$ is simply the time the chronoamperometric sequence is running in seconds. 
Our sensor contains an electrode surface area of 2.25 square centimeters and specifically tracks a target analyte with a reduction potential of 0.8 V according to Fick's law.
We utilize 1-Ethyl-3-methylimidazolium tetrafluoroborate ($[EMIM][BF_4$]) as the ionic liquid to act as a transducer over our electrodes.

Typically, a single chronoamperometric sequence operates anywhere between 6 and 60 seconds.
We borrow the method from \cite{france2025_fast_chronamp} to infer the analyte diffusion in order to speed up the sampling rate to 0.5-1 Hz.
Section \ref{sec:sm_fast_chronoamp} of the Supplementary Material presents more details on this technique.
Unlike our approach with the MOX sensors, we do adjust the fundamental operation of the EC sensors in Equation \ref{eq:cottrell}.

\subsection{Dataset}

We perform pattern recognition over the olfactory data via adaptive temporal difference learning.
We did not accumulate a dataset a priori for training the olfactory sensors, as we determined that adaptively learning the plume via real-time exploration approximation was sufficient, following many of the methods Sutton set forth in \cite{sutton1988learningtopredict}.
Data for the olfaction-vision models comes from \cite{france2025diffusiongraphneuralnetworks}, which fuses trusted and peer-reviewed olfactory \cite{goodscents, leffingwell, lee2023} and computer vision \cite{cocodataset} datasets, and is also the only known olfaction-vision dataset.

While modern reasoning models can associate from which objects an \textit{aroma} (e.g. lingual descriptors such as "fruity" and "musky") is coming in an image, others find they have difficulty in specifying from where a \textit{chemical compound} (e.g. $CO_2$, $CH_4$) is coming \cite{zhong2024sniffaispicyspicy, france2025diffusiongraphneuralnetworks, feng2025smellnetlargescaledatasetrealworld}.
In addition, leveraging a large vision-language reasoning model — most of which are billions or trillions of parameters in size — is neither practical nor necessary for real-time edge robotics.
To this end, we construct our own encoder to model olfactory-vision relationships, discussed in Section \ref{sec:ovnav}.

\subsection{Models}

\subsubsection{Olfaction-Only Navigation}
To track a plume, we must make some inference on its direction.
We use the time-delay estimation between the left and right sensors to inform the horizontal plume angle of attack against the sensor surface.
Let $x_L$ and $x_R$ represent the left and right lateral positions of the sensors onboard the aircraft, respectively.
For a planar odor front advecting with velocity $u$, wind speed $s$, and sensor separation $\Delta x = x_L - x_R$, we can define the expected sensor time constant $\hat{\tau}$ as:

\begin{equation}
    \hat{\tau} \approx -\frac{\Delta x \cdot u}{s^2}
\end{equation}

For a pure lateral baseline $\Delta x$ and known wind speed $s$, then the angle $\phi$ between the plume travel direction and longitudinal axis of the aircraft resolves to:

\begin{equation}
    \phi = \arcsin{\frac{\hat{\tau}s}{d}}
\end{equation}

\noindent This effectively converts a measured time lag into a heading command for the aircraft.
We clip $\phi$ such that $\phi \in [-90 \textdegree, -5\textdegree] \cup [5\textdegree, 90\textdegree]$ such that the UAV will hold the current heading for all $\phi \in (-5\textdegree, 5\textdegree)$.
Large angles can be likened to casting (exploratory) behaviors in insects, while smaller angles (meaning the UAV stays its current course) can be likened to surging (exploitive) behavior.

The UAV samples the readings every second from each of the two MOX sensors and every two seconds for the electrochemical sensors.
We apply dual-timescale exponential averaging to detect plume bouts while suppressing sensor drift and turbulence noise. 
This parallels filtering mechanisms observed in insect olfactory neurons.
We calculate this divergence $D$ by subtracting a long exponential moving average (EMA) with period $\beta$ from a short EMA with period $\alpha$, denoted by $E_\beta$ and $E_\alpha$, respectively:

\begin{equation}
    D = E_\alpha - E_\beta
\end{equation}

\noindent $D$ can be considered as a temporal-prediction-difference filter, which is mathematically adjacent to TD error computation in reinforcement learning, albeit not a learning mechanism.
Although $D$ does not constitute temporal-difference learning in the reinforcement-learning sense, it produces a temporal-prediction error — the difference between fast and slow odor expectations. 
This is conceptually similar to biological temporal difference filters that insect olfactory pathways use to emphasize odor onset and plume entry events.

We additionally compute a signal line $S$, which is a smoothed EMA of $D$ with period $\rho$, where $\alpha < \rho < \beta$. 
This signal line serves as a dynamic expectation of odor-trend behavior. 
Positive deviation between lines $D$ and $S$ indicates accelerating odor concentration, consistent with entry into the plume, whereas negative deviation reflects loss of odor momentum and triggers casting behavior. 
This dual-timescale architecture parallels both biological olfactory adaptation and momentum-based filtering methods in quantitative finance.


Formally, let $C(t)$ define a function over time $t$ for the olfaction sensor readings, where $C_L, C_R \in C$ define functions for the left and right sensors respectively.
Then

\begin{equation}
    D(t) = E_\alpha(C(t)) - E_\beta(C(t))
\end{equation}

\noindent And the signal line is

\begin{equation}
    S(t) = E_{\rho}(D(t))
\end{equation}

\noindent While $\phi$ gives the UAV the incident angle of the plume against the sensors, the bookkeeping of the $D$ and $S$ sequences allows the UAV to determine its proximity relative to the plume.
For the experiments shown here, we select empirically deduced values of 3, 8, and 5 for $\alpha$, $\beta$, and $\rho$, respectively.

\subsubsection{Olfaction-Vision Navigation}
\label{sec:ovnav}

\begin{figure}
  \centering
  \includegraphics[width=85mm]{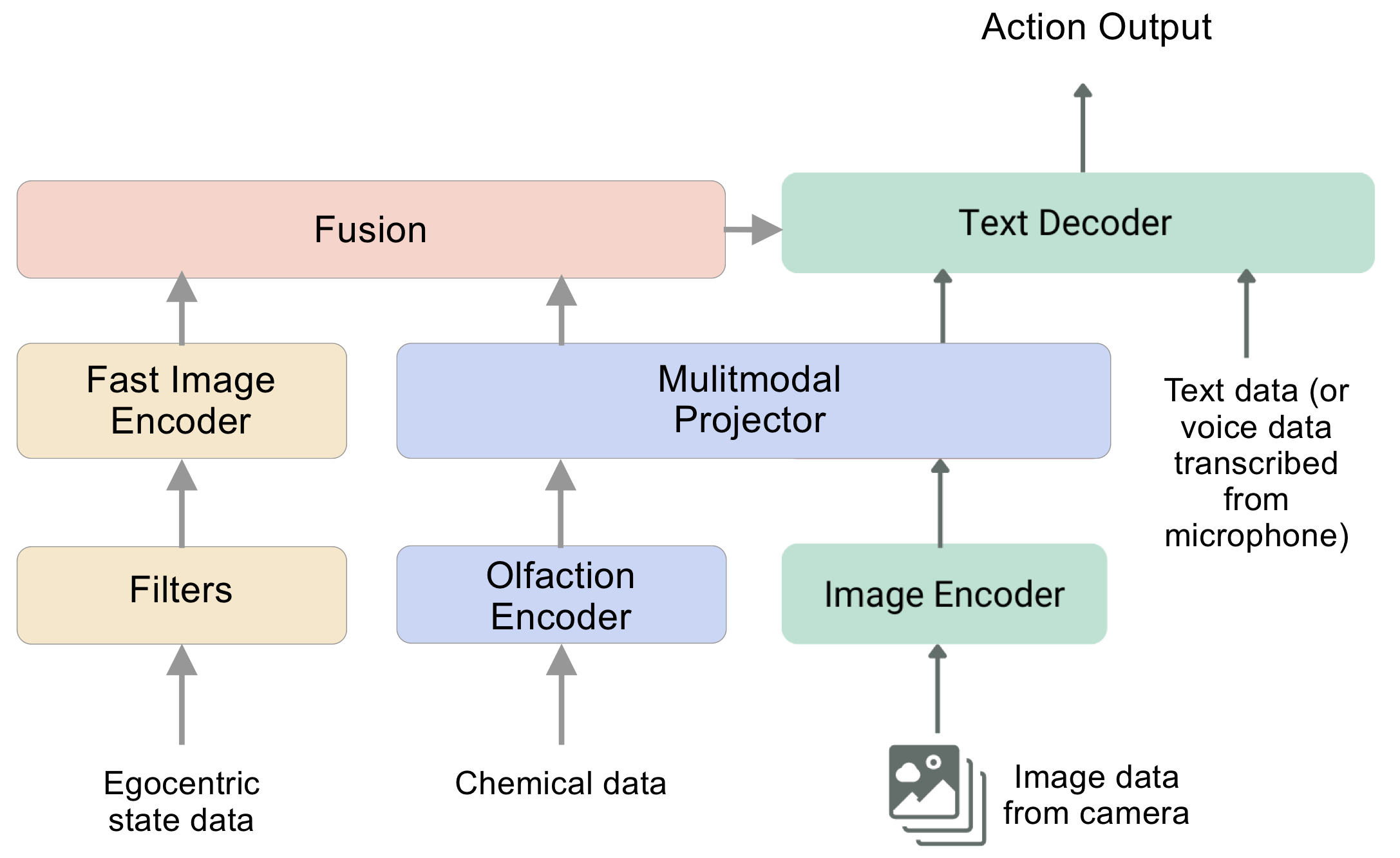}
  \caption{Olfaction-Vision Model Architecture. Blue and green boxes construct \textit{Component 1} based on \textit{COLIP}. The yellow boxes construct \textit{Component 2}. The output from both models inform reasoning for navigation.}
  \label{fig:ovlamArch}
\end{figure}

For navigation augmented by vision, we train and evaluate a multimodal machine learning model for understanding olfaction-vision relationships. 
Figure \ref{fig:ovlamArch} shows the main architecture.
It employs a CLIP-based vision encoder \cite{zhai2023siglip} to extract visual features from images. 
A separate 12-layer fully connected encoder processes chemical data from olfactory sensors, converting it into 138-dimension olfactory embeddings. 
We call this model \textit{COLIP (Contrastive Olfaction-Language-Image Pre-training)} since it builds on the original CLIP model.
A learned linear layer projects both visual and olfactory embeddings into a shared 512-dimension latent space, ensuring compatibility before further processing. 
A graph associator learns joint relationships between the projected visual and olfactory embeddings.
We evaluated two such models to assess tradeoffs in different architectures: a simple fully-connected network and a graph attention network (GAT).
Both models use the same 12-layer encoder.
The GAT captured more complex associations between the projected visual and olfactory embeddings, but we found difficulties in reliable performance when exporting to a format that could run at the edge on the UAV.
As a result, we do not leverage it for the final analysis because the extra computation and inference time required to run the model would contrast with our desire to run all computation fully at the edge.
Section \ref{sec:sm_ovm_details} in the Supplementary Material presents more details on the training parameters for these models.

Just like its predecessor, COLIP undergoes contrastive training using an InfoNCE (Information Noise Contrastive Estimation) loss to align olfactory and visual embeddings, encouraging the model to associate corresponding pairs closely in the latent space. 
These modality pairs encode together, effectively binding them in the process.
We optimize these encoders through the InfoNCE loss \cite{oord2019representationnceloss} shown below.

\begin{equation}
    L_{I,M} = - \log{\frac{\exp{( q_i^\intercal k_i / \tau )}}{\exp{(q_i^\intercal k_i / \tau)} + \sum_{j \neq i} \exp{(q_i^\intercal k_j / \tau)}}}
    \label{eq:nceloss}
\end{equation}

\noindent where $\tau$ denotes a scalar temperature that controls the smoothness of the softmax distribution and $j$ defines "negatives", or unrelated observations.
We presume each example $j \neq i$ in the mini-batch to be a negative.
The loss makes the embeddings $q_i$ and $k_i$ closer in Euclidean distance in joint embedding space.
This consequently aligns $I$ with $M$ conveniently for joint learning.
A binary classification loss from the graph head learns relationships between modalities, distinguishing between matching and non-matching olfaction-vision pairs.

Many approaches to robotic control struggle with a core tradeoff: backbones of large vision models are highly general but too slow for real-time operation, while traditional robot visuomotor policies are fast but lack generalization \cite{eschmann2023learning}. 
We resolve this tradeoff by introducing a dual-system architecture. 
Component 1 is the OVM that runs at 0.1-0.2 Hz. 
It handles high-level scene understanding and multimodal comprehension, enabling broad generalization across objects, environments, and goals.
Component 2 is a high-speed reactive vision-olfaction module containing a fine-tuned YOLOv11 \cite{khanam2024yolov11overviewkeyarchitectural} for rapid vision inference running inference at 1-2 Hz.
Component 1 mainly confirms whether the acceptance criteria for locating the odor source have been achieved (see Section \ref{sec:acceptance} for more on this).
Component 2 synchronizes at the same sampling rate as the olfaction sensors, and helps convert the latent representations from the OVM into precise, continuous UAV actions.

This decoupled design lets each component function at its optimal temporal scale. 
Component 2 can deliberate slowly over abstract goals, while Component 1 executes rapid, low-latency commands in response to dynamic conditions. 
This relieves the need to model increased complexity in our plume models and simulations.
Furthermore, the separation of both components allows for independent development and iteration.

\begin{figure}
  \centering
  \includegraphics[width=75mm]{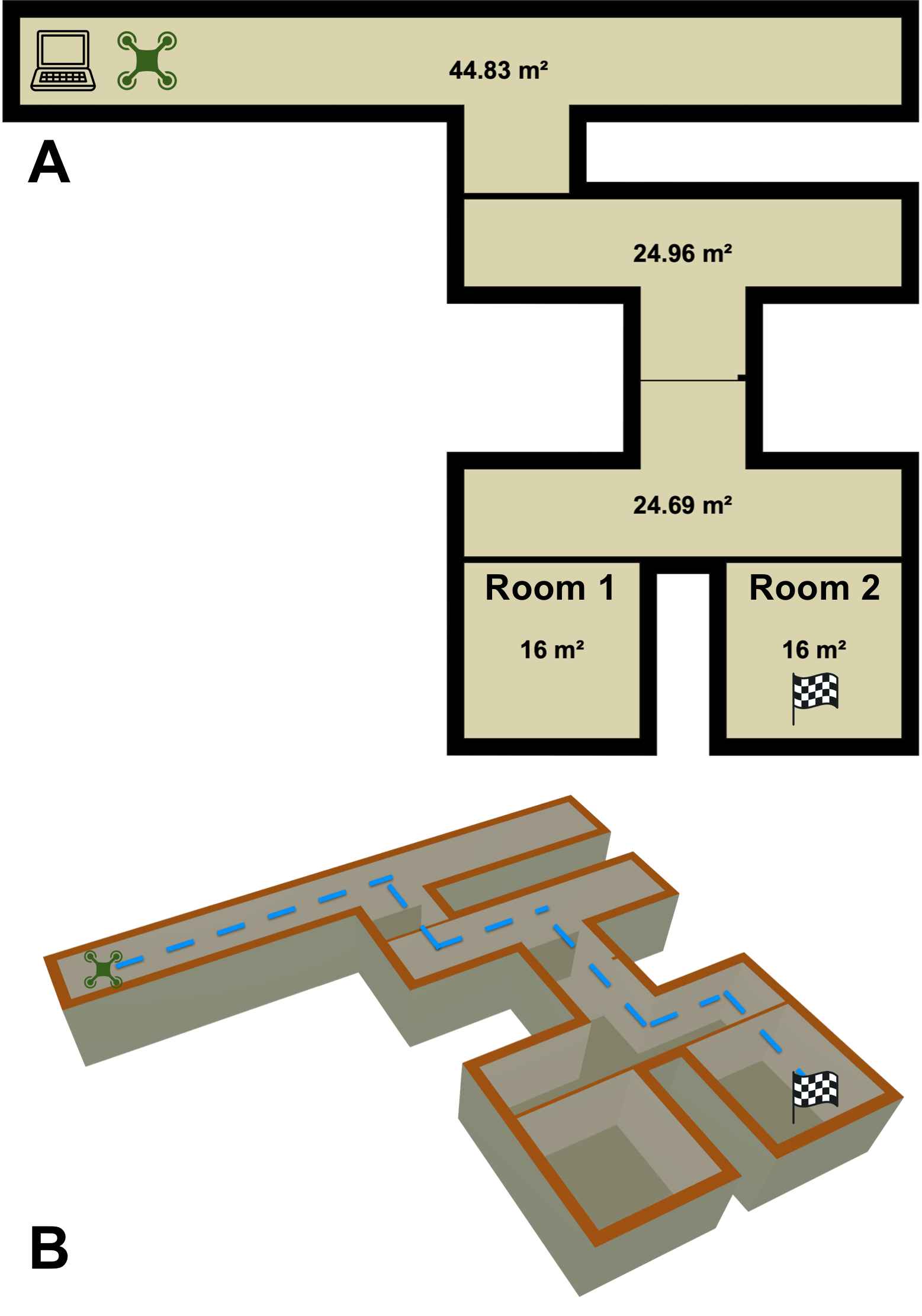}
  \caption{Illustration of the course developed for olfactory navigation by the UAV. Figure A shows a bird's-eye-view of the course, where the ground station (laptop icon) is located at the top left and the UAV (quadcopter icon) takes off next to it; Room 2 hosts the target compound and the plume is formed from here. Figure B shows an isometric view of the ideal path the UAV should pursue to locate the source.}
  \label{fig:course}
\end{figure}

\subsection{Navigation \& Plume Models}

Plume models are inherently complex and require advanced computational fluid dynamics engines to achieve reasonable approximations. 
We seek a set of algorithms that generalize well to all chemical compounds and follow simple rules based on environmental evidence, not unlike those insects follow \cite{Singh2020, Singh2023, Crimaldi2022DynamicSensing}.
The primary navigation model was a simple bout detection algorithm \cite{schmuker16-bout-detection, sniffybug-swarm-duisterhof21} in which first and second derivatives align with short and long moving averages of the olfactory signal coupled with inertial data to determine the UAV actions.
For brevity, we denote this model as \textit{olfactory inertial odometry} (\textit{OIO}), building off the principles from \cite{france_2025_oio_method}.
We model plumes with simple Gaussian time-series processes, inserting blanks \cite{dennler_2025_neuromorophicandolfaction} into the data to resemble wind shifts and plume volatility.
The robot has seven actions available depending on temporal plume dynamics: surge forward, cast in one of four directions, pause, or land.
Selecting the "land" action indicates that the UAV is confident it has found the plume source.
Our perception reasoning is highly susceptible to hysteresis due to the ability of shifting plume dynamics.
We employ tuned Kalman filters and sensor synchronization to counter this.
We implemented a custom plume environment using the \textit{Gymnasium} framework \cite{Brockman2016:DBLP:journals/corr/BrockmanCPSSTZ16} for additional simulations in plume tracking.
We designed factors such as temperature, relative humidity, barometric pressure, wind direction, wind magnitude, wind sparsity, and air chemical composition to all be tunable hyperparameters of the plume.
The design follows Gaussian plume principles \cite{Crimaldi2022DynamicSensing} and a Dryden turbulence model \cite{dryden1952}.
Section \ref{sec:sm_plume_env} of the supplementary material presents more details about this plume environment and its construction.

We implemented a second navigation model based on reinforcement learning (RL) to provide a more comprehensive assessment of the navigation.
\textit{Q-learning} \cite{Watkins1989} is a standard temporal difference RL algorithm that many applications use.
Due to the dynamism of plumes and the UAV's frequent probability to travel off plume, we employ eligibility traces to harmonize immediate and long-term rewards. 
$TD(\lambda)$ algorithms leverage eligibility traces to give credit to more recent $\lambda$ states instead of only the current state ($\lambda = 0$).
Consequently, we leverage the \textit{TD($\lambda$)} derivatives of \textit{Q}-learning and Expected SARSA, called $Q(\lambda)$ and \textit{Expected SARSA($\lambda$)}, respectively \cite{sutton2018reinforcement}.
While we trained both models in simulation, we found Expected SARSA $(\lambda)$ to be the model most robust to reasoning about blank pockets of air, and thus selected it as a finalist for evaluation in our experiments.
Expected SARSA$(\lambda)$ proved most robust for air with continuous gradients, but this scenario is not practical for real world applications.
Section \ref{sec:sm_rlalgo} of the Supplementary Material presents exact training parameters for these models.

With either model, we discretize the aircraft's movements such that it can only move 10 \emph{cm/sec} or $\frac{\pi}{2}$ \emph{radians/sec} at each time step.
In essence, we give identical linear and angular acceleration commands for every command. 
We followed a policy that allows the UAV to move and wait for 2 seconds to sample continually until done. 
For each ``move'' command, the magnitude of acceleration and velocity is identical to provide a level of experimental control and reproducibility. 
The consistency here also ensures we did not encounter further nonlinearities induced by the payload bias. 
We find that making the actions continuous was not necessary, as doing so added more complexity to our simulations and did not substantially improve them. 
In addition, the infrared sensors enable the UAV to make a decision on how to move forward.
We acknowledge that enabling continuous action spaces could present an opportunity for future work.

\subsection{Course}

Finally, we transfer our solution to the real world with flight tests on the UAV.
We deploy our UAV over a 200 $m^2$ course, depicted in Figure \ref{fig:course}.
We start by placing 100 mL of ethanol within a home diffuser in Room 2 as the checkered flag indicates. 
The diffuser diffuses the ethanol at a rate of 2.16 mL per minute. 
We place the diffuser near an 80-cm home fan that exhausts air at 1415 liters per minute in order to develop a small plume for traceability of the compound.
The robot starts at the far adjacent end of the map as the UAV icon in the figure indicates.
We observe 5 runs each with the target compound lying in both rooms 1 and 2.
At the beginning of each run, we give the compound 5 minutes to disperse and form the start of a plume before we command the UAV for takeoff. 
After takeoff initiates, no human intervenes with the UAV until it sends a response back to the ground station that it has found the source of the target compound.
Between each run, we vent the course and take air measurements to ensure that the next iteration may begin with a clean baseline and to encourage as reproducible results as possible.
Due to the length of time needed to properly vent the course, the uncontrollable HVAC schedules within the test building, and other experimental controls needed to re-initialize to a stable environment, the number of statistical runs we can perform is limited.
As a result, for each of the four tasks, we analyze five trials.
We note that prior work from \cite{sniffybug-single-burgues19} and \cite{sniffybug-swarm-duisterhof21} has a low count of statistical runs (identical to ours), which we suspect these factors also caused.

\begin{figure}
  \centering
  \includegraphics[width=75mm]{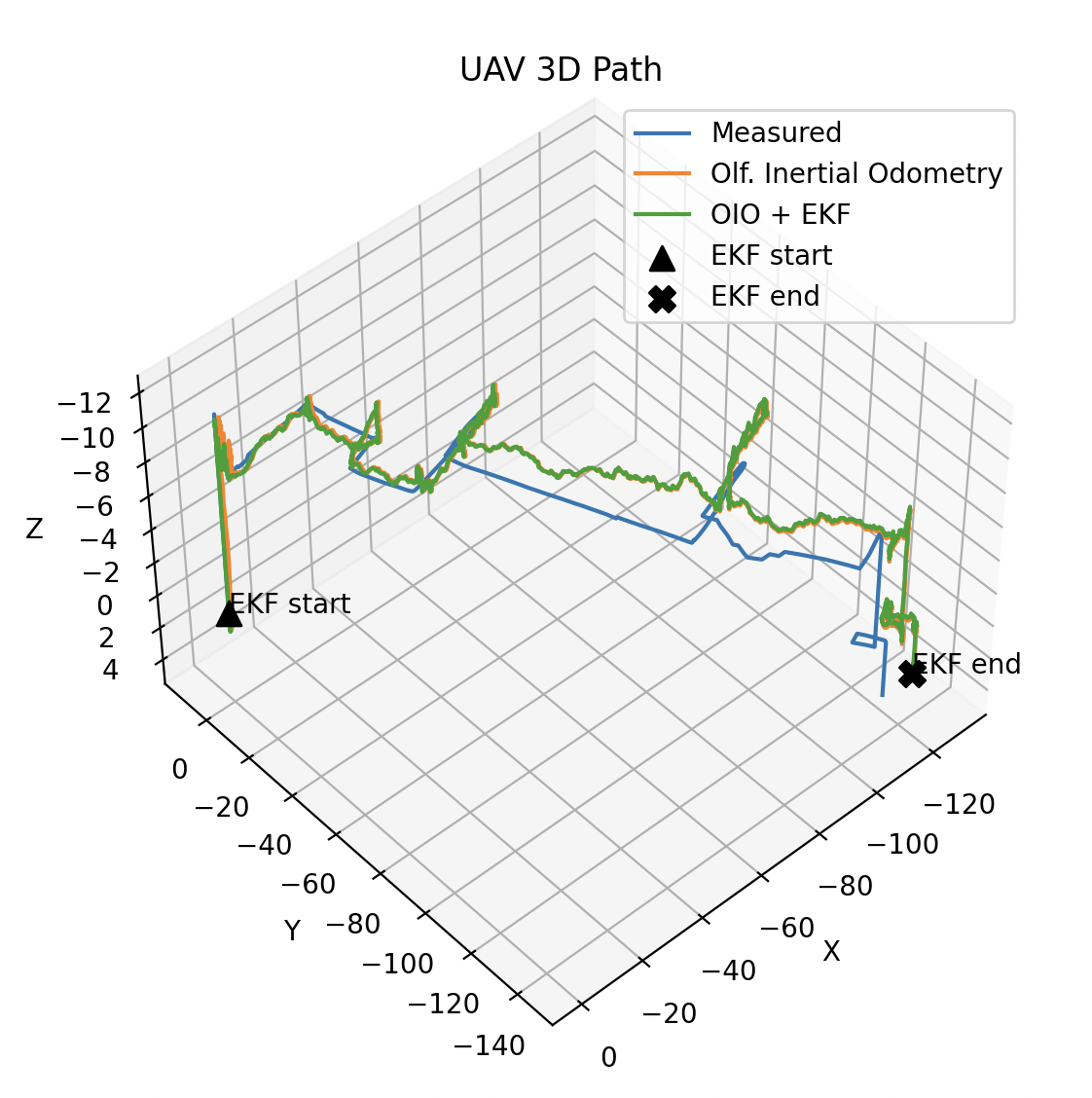}
  \caption{The IMU replay of the UAV traveling over the whole course. Unfiltered olfactory inertial odometry (OIO) \cite{france_2025_oio_method, france_2025_oio_cal} along with filtered OIO via EKF appear alongside naked measurements from the IMU. The addition of the olfaction sensors and OPU as payload significantly biased the IMU. This made compound mapping via long-term inertial odometry nearly impossible and drove our decision for the task termination criteria in Section \ref{sec:acceptance}. Measurements are in centimeters.}
  \label{fig:uav_est_pos}
\end{figure}

\begin{table*}
    \caption{Final Results}
    \begin{center}
    \begin{tabular}{|c|c|c|c|c|c|}
    \hline
    \textbf{Method}& \textbf{Sensor Type} & \textbf{Nav. Algorithm} & \textbf{Average Time $\mu_t$ (s)}& \textbf{Sigma Time, $\sigma_t$ (s)}& \textbf{Best Time, $\beta_t$ (s)} \\
    \hline
    Olfaction& MOX& OIO& 98.38& 14.84& 71.59\\
    Olfaction& EC& OIO& 112.22& 16.70& 97.65\\
    Olfaction& MOX& E.SARSA$(\lambda)$& 103.99& 14.89& 82.73\\
    Olfaction& EC& E.SARSA$(\lambda)$& 121.85& 16.77& 97.10\\
    Olfaction + Vision& MOX& OIO&  94.03& 14.68& 66.12\\
    Olfaction + Vision& EC& OIO& 107.71& 17.07& 80.50\\
    \hline
    \end{tabular}
    \label{tab:resultsTable}
    \end{center}
\end{table*}

\subsection{Task Termination Criteria}
\label{sec:acceptance}
One defining contribution of our work is how a UAV decides it is "done".
For the robot to determine whether it has found the source of the gas, it must evaluate a hypothesis about whether it believes it has found the highest concentration of the plume given its observed samples.
On our selected UAV, the IMU bias from the added payload was significant, making long-term positional estimation very difficult.
The UAV could perform all maneuvers commanded, but the added payload dampened the magnitude of each command.
We found this dampening factor non-linear and partially dependent on battery life.
This made the error of the commands difficult to model, but we found short-term estimation more practical to filter via extended Kalman filters and reasoned via OIO accordingly.
See Figure \ref{fig:uav_est_pos} as an example of the IMU replay that occurred over the map detailed in Figure \ref{fig:course}.

It is concluded that the UAV reaches the gas source when it lands within 1 meter of the diffuser.
To evaluate when the UAV has reached the gas source, we construct a minimum-variance unbiased point estimator $\hat{C}$ to approximate the true highest concentration of the gas within the environment, given by: 

\begin{equation}
    \hat{C} = m\frac{k}{k+1} - 1
    \label{eq:unbiasedpointestimate}
\end{equation}

\noindent where $m$ is the largest concentration the olfaction sensors observed and $k$ is the total number of samples observed.
We assume that once a chemical packet has been observed, it can be observable again, as in drawing with replacement \cite{clark2021lessonsgermantankproblem, Simon2024_germantankproblem}.

The point estimate becomes reliable only after long sessions with several observed samples.
However, because we desire a generalized solution to both short and long exploration sessions, we compute a confidence interval for $\hat{C}$.
We compute the $p$-th and $q$-th quantiles of the sample maximum $m$ to form the interval $[\hat{C}q^{1/k}, \hat{C}p^{1/k}$], which yields the corresponding confidence interval for the maximum estimated gas source concentration over the full population of gas samples:

\begin{equation}
    \left[\frac{m}{q^{1/k}}, \frac{m}{p^{1/k}}\right]
    \label{eq:confidenceinterval}
\end{equation} 

As an example, assume the UAV has observed 20 samples and believes that it has found the source of the gas with an observed final concentration $m$.
The point estimate for $k=20$ samples would be $1.05m$ based on Equation \ref{eq:unbiasedpointestimate}.
By Equation \ref{eq:confidenceinterval}, taking a 95\% confidence interval would return 

\begin{equation}
    [\frac{m}{0.975^{1/20}}, \frac{m}{0.025^{1/20}}] = [1.001m, 1.2025m] \approx [m, 1.2m]
\end{equation}

\noindent This indicates that, given the observed plume concentrations, we can be 95\% confident that the true highest concentration of the plume lies between the current observed highest concentration $m$ and a concentration 20\% larger.
From here, the UAV can determine whether to keep navigating or to end.
In general, we strive to stay within 25\% of the expected maximum odor concentration based on a 95\% confidence interval; so in this case, the UAV would land and stop navigating.


\section{Results}

Table \ref{tab:resultsTable} shows final results for our experiments.
For all trials, the UAV successfully found the source of the target compound, which was our minimum bar for success.
We observe that electrochemical-based olfactory sensors require more time to complete the task in general, and we suspect this is due to the fact that EC sensors are more sensitive to perturbations in the target compound's concentration, which could cause them to overfit to noise or small eddy currents in the plume.
For both MOX and EC sensors, we note a small bump in performance from adding vision as a modality by which to localize the odor.
In all tasks, the UAV only uses olfaction to navigate.
OIO showed more consistency in navigation times than RL.
We suspect that this is due to RL over-optimizing the navigation path by attempting to predict ahead or weighing states too far in the past for future decisions due to the eligibility traces.
In future work, we expect that wrapping OIO with RL will become more competitive (or objectively better) than naked OIO with further tuning.
In the \textit{Olfaction + Vision} tasks, the UAV only activates the camera when it can no longer maximize the olfactory gradient, which usually occurs when the UAV is confident it has reached the target compound based on the termination criteria.
When the UAV uses only olfaction, we observed that it would spend several seconds slowly moving around the chemical source, trying to maximize the olfactory signal.
Because diffusion of the gas is more uniform the closer one is to the gas's source, an apparent gradient may not be easy to determine, and the UAV will waste several seconds moving around very close to the source.
This is where vision became additive: using a visual confirmation that the UAV has found the source allows the UAV to not waste time and battery chasing small gradients.
As a consequence, vision shortened navigation times by shaving a few seconds off the end of each trial, where olfaction-only navigation would have kept the UAV flying.

\section{Discussion \& Limitations}
\textbf{The most important conclusion we drew from our experiments was that olfaction is enough to navigate when localizing to an odor source}.
We believe this is the first documented attempt to enable a robot to navigate purely by scent without any planning, a priori knowledge of the environment, or location aids such as UWB radio tags.
We attempt to show our results generalize by performing experiments with two different olfaction sensor types.
However, we acknowledge that a larger variety of chemical sensors exists.
In addition, we perform our work here over only one chemical compound: ethanol.
This is intentional in order to build off of the assumptions and work by Burgues, et al. \cite{sniffybug-single-burgues19} and Duisterhof, et al. \cite{sniffybug-swarm-duisterhof21}, as they demonstrate olfactory navigation with radio tags and metal oxide sensors in tracking ethanol.
With our work here, we attempt to establish a generalizable framework that builds off the accepted state of the art, and that can enable intuitive extrapolation of our hardware and algorithms to other compounds with little or no tuning.
We are performing ongoing work to support this and demonstrate said framework on a variety of gases in outdoor conditions.

We note that our results show that the UAV localized itself to the chemical source on \textit{every} trial, while the work of Shigaki, et al. \cite{shigaki2018quadcopter} noted a non-perfect localization rate.
We are also performing ongoing work to demonstrate the generalization of our framework over other environments, chemical compounds, and sensor types such as graphene-based sensors, spectroscopy sensors, and photo-ionization detectors.
We are also performing ongoing work to evaluate our framework over wider kinematic envelopes.
We acknowledge that exact replication of any olfactory navigation task is difficult due to the inherent nonlinearities of gas dynamics and the difficulty in replicating exact initial chemistry conditions.
As a result, the focus of our work here is to establish a simple, reproducible, and generalized framework on which other researchers can build to conduct additional experiments that lead to further improvements and increased reproducibility.

\section{Conclusion}
\label{sec:conclude}
Scent-based navigation is a very nuanced robotics task.
Organisms that leverage olfaction to perceive the world optimize over various criteria, yet none of them are standardized.
With our framework and experiments here, we show that, to localize a UAV to a chemical source, olfactory perception is enough.
We incorporate many nature-inspired principles to define methods for the first instance of stereo olfaction-vision odor localization and demonstrate their practicality with real aerial robots.
We make our code, simulations, hardware schematics, and datasets available to the community to encourage advancement in the field.
We hope the techniques presented here encourage more work in olfactory robotics and bring us one step closer to enabling the sense of smell for machines.

\bibliography{neurips_2025}
\bibliographystyle{IEEEtran}

\newpage
\setcounter{page}{1}
\onecolumn
\section*{Supplementary Material}

\subsection{Overview of Related Work}
\label{sec:sm_relatedwork}
We present a high-level comparison of our work to previous research in olfaction navigation.
We strove to maintain parity with existing research and build off of their assumptions and methods for a level of experimental control while integrating our own optimizations.
These papers are the closest we found to our research.

\begin{center}
\begin{longtable}{>{\raggedright\arraybackslash}p{2cm} >{\raggedright\arraybackslash}p{3cm} >{\raggedright\arraybackslash}p{3cm} >{\raggedright\arraybackslash}p{3cm} >{\raggedright\arraybackslash}p{5cm}}
\caption{Comparison to Relevant Prior Works} \\
\toprule
Aspect & Shigaki et al. (2018/2023 variants) & Burgués et al. (2019) & Duisterhof et al. (2021) & Our Work \\
\midrule
\endfirsthead
\toprule
Aspect & Shigaki et al. (2018/2023 variants) & Burgués et al. (2019) & Duisterhof et al. (2021) & Our Work \\
\midrule
\endhead
\bottomrule
\endfoot
Robot Type & Pocket-sized quadcopter (2018) or ground crawler (2023); single unit. & Nano quadcopter (Crazyflie). & Swarm of nano quadcopters (Crazyflie). & Modified DJI Tello UAV; single unit but with custom harness for stereo sensors. Larger payload for dual sensors and vision integration. \\
Sensors & Single MOX sensor (2018) or dual alcohol + wind sensors (2023). & Single MOX (TGS 8100). & Single MOX per drone + laser rangers. & Dual MOX/EC sensors in stereo (antennae-like); IR for obstacles; vision complementary. First stereo olfaction demo; multiple sensor types for generalization. \\
Navigation Approach & Bio-inspired (surge-cast/zigzag); real-time with odor frequency. & Predefined sweeping path; bout detection or concentration mapping. & Mapless bug algorithm with PSO; real-time swarming. & Adaptive TD learning (OIO/Q($\lambda$)); real-time plume modeling with temporal differences. TD($\lambda$) for plumes; no a priori path; vision for "last-mile" confirmation. \\
Course Size & Small (~1.5-2m start distance; indoor/outdoor open areas). & 160m² indoor. & 10x10m (100m²) with obstacles. & 200m² indoor with rooms. Largest course; more complex (multi-room plume dynamics). \\
Success Rate & ~80-90\% (2018: non-perfect; 2023: high but variable in complex envs). & N/A (mapping error 1-2m; not framed as success rate). & 92\% in sim; 11/12 real runs. & 100\% across 20+ trials. Perfect localization; robust to hysteresis without overfitting. \\
Real-Time vs. Post-Mapping & Real-time (autonomous onboard). & Post-mapping (builds 3D map after flight; potential for real-time discussed but not implemented). & Real-time (mapless, autonomous). & Real-time (edge-only; no post-analysis). True real-time source localization (not just mapping like Burgués); straight to source. \\
Positioning Aids & None (onboard only). & External UWB anchors for 3D positioning. & Onboard UWB for relative localization. & None (only onboard IR + olfaction; naive to environment). Fully aid-free; no UWB/radio tags like Burgués/Duisterhof. \\
Other & Focus on bio-mimicry; ethanol source. & Gas mapping; ethanol. & Swarm coordination; sim-to-real. & Olfaction-vision models; custom hardware; open source. Vision augmentation; edge ML; sim plume env. \\
\bottomrule
\end{longtable}
\end{center}

\FloatBarrier
\subsection{Control Algorithms}
\label{sec:sm_autopilot}
We based the autopilot control loop on a \textit{Proportional-Integral Controller with Rate Feedback}, or \textit{PIR} for short.
Figure \ref{fig:autopilot_diagram} below shows the control diagram.
Table \ref{tab:autopilot_table} explains each parameter.
Table \ref{tab:autopilot_gains} shows the values for each gain of each controller.

\begin{figure}
  \centering
  \includegraphics[width=160mm]{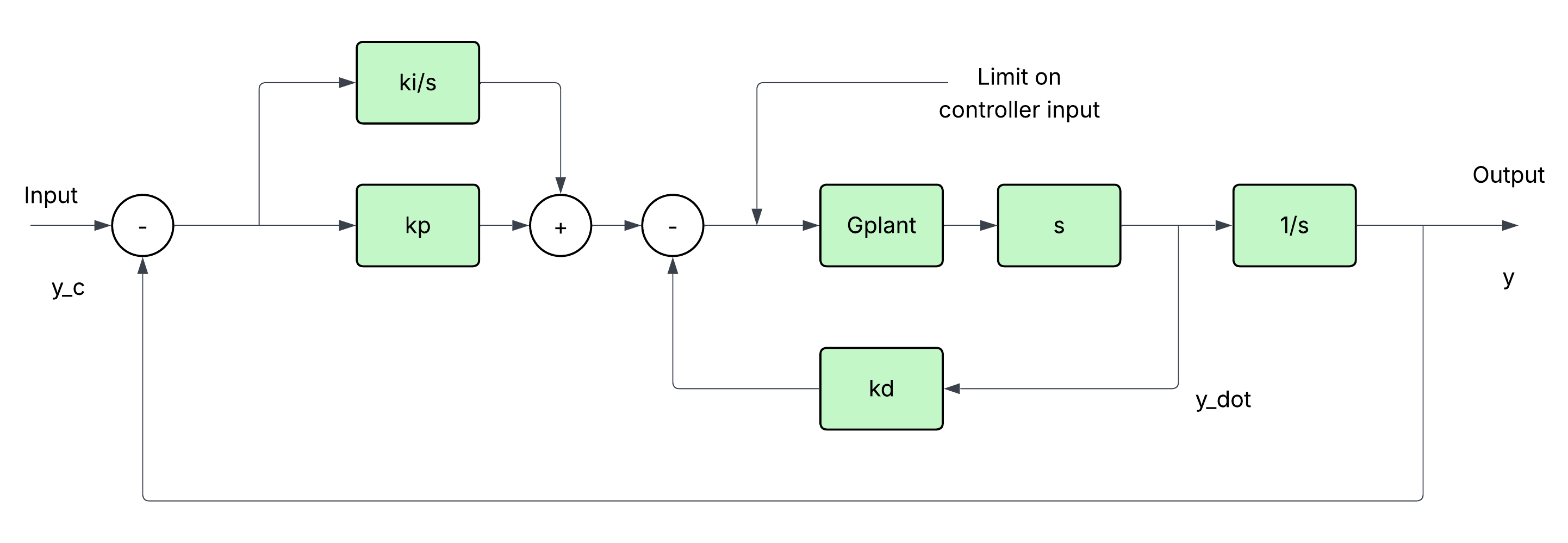}
  \caption{We designed the control loop to model the DJI Tello control algorithms and understand how to sequence the olfaction sensors with the other sensors.}
  \label{fig:autopilot_diagram}
\end{figure}

\begin{table}[htb]
    \caption{PI Controller with Rate Feedback}
    \begin{center}
    \begin{tabular}{|c|c|}
    \hline
    \textbf{Parameter}& \textbf{Definition}\\
    \hline
    $G_{plant}$& Controller\\
    $y_c$& Closed loop command\\
    $y$& Current system response\\ 
    $\cdot{y}$& Rate feedback\\
    $dt$& timestep, in seconds\\
    $k_i$& integral gain\\
    $k_p$& proportional gain\\
    $k_d$& derivative gain\\
    \hline
    \end{tabular}
    \label{tab:autopilot_table}
    \end{center}
\end{table}

\FloatBarrier
\begin{table}[htb]
    \caption{PID gains for each PIR controller}
    \begin{center}
    \begin{tabular}{|c|c|}
    \hline
    \textbf{Parameter}& \textbf{Value}\\
    \hline
    $k_p$ (roll)& 0.010\\
    $k_i$ (roll)& 0.002\\
    $k_d$ (roll) & 0.006\\
    $k_p$ (altitude)& 0.070\\
    $k_i$ (altitude)& 0.015\\
    $k_d$ (altitude)& 0.060-0.080\\
    $k_p$ (pitch)& 0.010-0.012\\
    $k_i$ (pitch)& 0.002\\
    $k_d$ (pitch)& 0.010\\
    $k_p$ (yaw)& 0.100-0.200\\
    $k_i$ (yaw)& 0.012\\
    $k_d$ (yaw)& 0.100\\
    $k_p$ (velocity lateral)& 0.100\\
    $k_i$ (velocity lateral)& 0.050\\
    $k_d$ (velocity lateral)& 0.000-0.010\\
    $k_p$ (velocity longitudinal)& 0.100\\
    $k_i$ (velocity longitudinal)& 0.050\\
    $k_d$ (velocity longitudinal)& 0.000-0.010\\
    \hline
    \end{tabular}
    \label{tab:autopilot_gains}
    \end{center}
\end{table}


\subsection{Inference Technique for Electrochemical Sensors}
\label{sec:sm_fast_chronoamp}
\begin{figure}
  \centering
  \includegraphics[width=150mm]{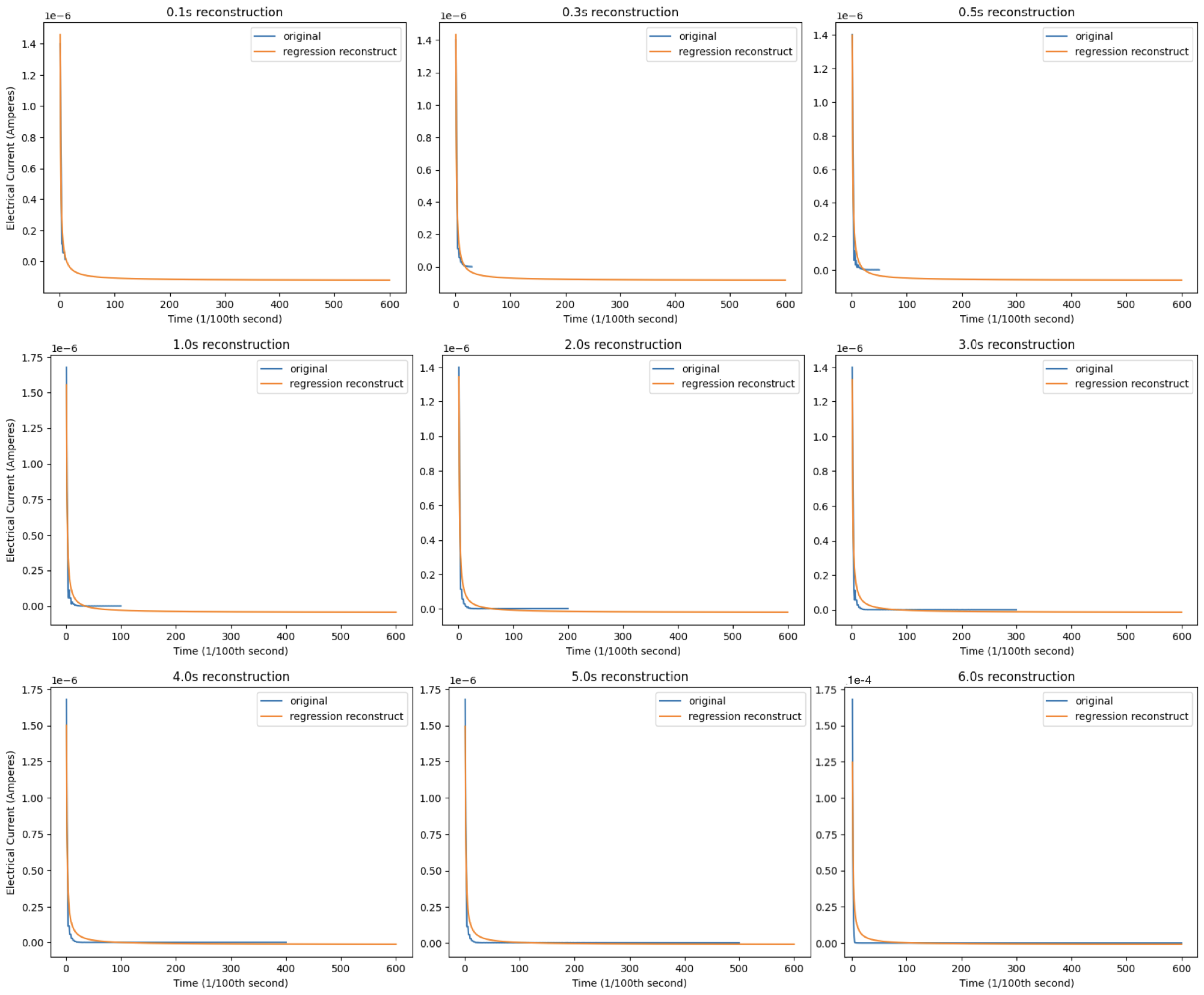}
  \caption{Prediction of the full 6-second diffusion current for chronoamperometry over various time periods at a 100-Hz measurement frequency.}
  \label{fig:sm_fast_chronoamp_100hz}
\end{figure}

\begin{figure}
  \centering
  \includegraphics[width=180mm]{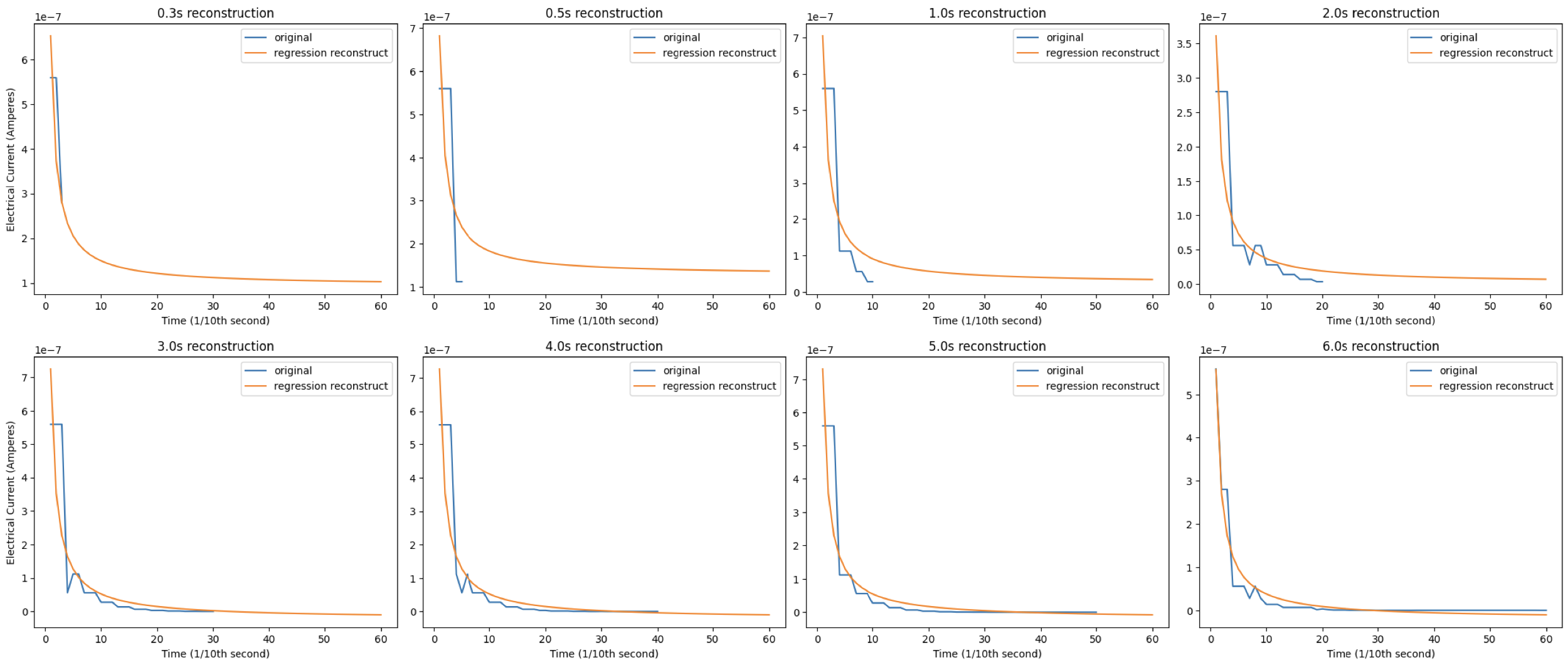}
  \caption{Prediction of the full 6-second diffusion current for chronoamperometry over various time periods at a 10-Hz measurement frequency.}
  \label{fig:sm_fast_chronoamp_10hz}
\end{figure}

For metal oxide sensor measurement, we use principles from Dennler, et al. in \cite{dennler_high-speed_2024}.
For measurements with electrochemical sensors, we borrow the fast inference technique for chronoamperometry from \cite{france2025_fast_chronamp}.
We evaluate two sampling rates for chronoamperometry: 10 Hz and 100 Hz.
We determined that, while 10 Hz was computationally cheaper, 100 Hz gave us more points from which the fast inference algorithm can infer, so we set 100 Hz as the sampling rate for deployment on the UAV and cut the chronoamperometry measurement after 1.0 second.
Figure \ref{fig:sm_fast_chronoamp_100hz} shows plots for assessing how well the inference technique predicted the full 6-second diffusion sequence for a 100-Hz measurement frequency, and Figure \ref{fig:sm_fast_chronoamp_10hz} shows the same for a 10-Hz measurement frequency.

\FloatBarrier
\subsection{Training of Olfaction-Vision Models}
\label{sec:sm_ovm_details}

We trained an encoder to compress olfactory data into a format that could jointly learn with CLIP through a graph associator model.
We intentionally kept this encoder simple in architecture due to issues in exporting advanced layers that we observed with the graph associator models (more on this below).
The encoder is a very deep model, but we intentionally over-parameterized it due to the low additional computational overhead and because there is a general lack of open-sourced olfactory training data from which to learn (depth and over-parameterization have shown to help models generalize in the absence of plentiful data \cite{Goodfellow_2016}).
Details on encoder training appear below:

\begin{table}[htb]
    \caption{Olfaction Encoder}
    \begin{center}
    \begin{tabular}{|c|c|}
    \hline
    \textbf{Parameter}& \textbf{Value}\\
    \hline
    layers& 12\\
    layer types& Fully connected only\\ 
    input dimension& 138\\
    output dimension& 512\\
    batch size& 16\\
    learning rate& 1e-4\\
    activation functions& ReLU\\
    epochs& 1100\\
    loss function& Information noise-contrastive estimation\\
    loss function temperature& 0.07\\
    optimizer& Adam\\
    \hline
    \end{tabular}
    \label{tab:sm_olf_encoder_params_table}
    \end{center}
\end{table}

\FloatBarrier
We trained two graph associator machine learning models to learn olfaction-vision relationships: a fully-connected (FC) neural network and a graph-attention (GAT) neural network.
We optimized the simple FC network to export for use at the edge without external API calls.
This model has lower computational demand, lower layer complexity (and thus easier exportability), and faster inference.
Details on its architecture appear below:

\begin{table}[htb]
    \caption{FC Graph Associator}
    \begin{center}
    \begin{tabular}{|c|c|}
    \hline
    \textbf{Parameter}& \textbf{Value}\\
    \hline
    layers& 4\\
    layer types& Fully connected only\\ 
    input dimension& 512\\
    output dimension& 512\\
    embedding dimension& 512\\
    batch size& 16\\
    learning rate& 1e-4\\
    activation functions& ReLU\\
    epochs& 1100\\
    loss function& Information noise-contrastive estimation\\
    loss function temperature& 0.07\\
    optimizer& Adam\\
    \hline
    \end{tabular}
    \label{tab:sm_fc_params_table}
    \end{center}
\end{table}

\FloatBarrier
The GAT associator learned more complex relationships between CLIP and olfactory representations.
However, we noted a lack of consistency in inference performance between what we observed when training the models and what we actually observed when exported to the UAV firmware.
As a result, we did not move forward with this architecture as the baseline model.
Details on GAT training appear below:

\begin{table}[htb]
    \caption{GAT Graph Associator}
    \begin{center}
    \begin{tabular}{|c|c|}
    \hline
    \textbf{Parameter}& \textbf{Value}\\
    \hline
    layers& 4\\
    layer types& 2 fully connected, 2 GAT Convolution\\ 
    input dimension& 512\\
    output dimension& 512\\
    embedding dimension& 512\\
    batch size& 16\\
    learning rate& 1e-4\\
    activation functions& ReLU\\
    epochs& 1100\\
    loss function& Information noise-contrastive estimation\\
    loss function temperature& 0.07\\
    optimizer& Adam\\
    \hline
    \end{tabular}
    \label{tab:sm_gat_params_table}
    \end{center}
\end{table}

\FloatBarrier
Figure \ref{fig:ovm_results} below shows the loss plot of the final training run for both graph associator models jointly trained with the olfactory encoder + CLIP.

\begin{figure}[htb]
  \centering
  \includegraphics[width=110mm]{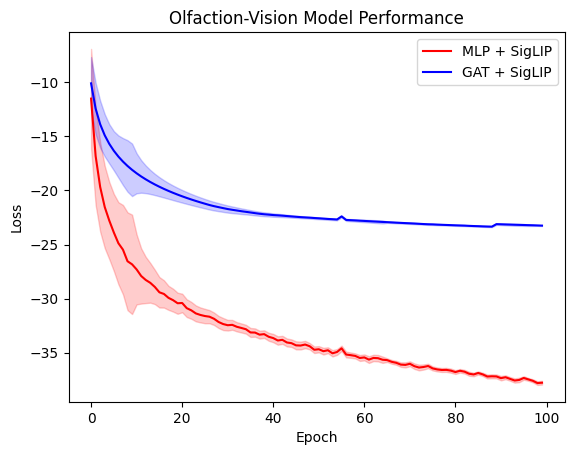}
  \caption{Training losses for both olfaction-vision models. Note that the GAT-based model is more accurate, but requires more compute and inference time, and thus violates our constraints for full edge computation. As a result, we use the MLP-based model for our final results.}
  \label{fig:ovm_results}
\end{figure}

References \cite{france_2025_ovlm_data_hf} and \cite{france_2025_ovlm_models_classifiers_hf} respectively provide full code to reproduce the training data and results for both models.

\FloatBarrier
\subsection{Training of Reinforcement Learning Algorithms}
\label{sec:sm_rlalgo}

In order to construct any learning algorithm for navigation, we needed to solidify the states for which the UAV could possibly reside and the actions available to it in each state.
We deduced that for each of 9 different states, 7 actions were possible: turn slightly left (45°), turn slightly right (45°), turn hard left (90°), turn hard right (90°), surge forward, cast left, or cast right.
Table \ref{tab:sm_rl_states_actions} below details the 9 states.

\begin{table}[htb]
    \caption{State Table for Each Navigation Agent}
    \begin{center}
    \begin{tabular}{|c|c|c|c|}
    \hline
    \textbf{-} & \textbf{Off-Plume} & \textbf{On-Plume (Low Concentration)} & \textbf{On-Plume (High Concentration)}\\
    \hline
    \textbf{Left Sensor $<$ Right Sensor}&0&3&6\\
    \textbf{Left Sensor == Right Sensor}&1&4&7\\
    \textbf{Left Sensor $>$ Right Sensor}&2&5&8\\
    \hline
    \end{tabular}
    \label{tab:sm_rl_states_actions}
    \end{center}
\end{table}

We selected two TD($\lambda$) algorithms for simulation as a possible navigation algorithm: Q($\lambda$) and Expected SARSA($\lambda$).

Q-learning is an off-policy temporal-difference (TD) method that directly estimates the optimal action–value function without requiring adherence to the policy that drives exploration \cite{Watkins1989}.
It learns a policy that selects state–action transitions yielding the highest reward and ranks among the most widely adopted TD algorithms in reinforcement learning due to its computational simplicity and its effectiveness in approximating long-term cumulative return.

Greedy Q-learning performs well in settings where rewards are clearly defined for each state.
From an intuitive standpoint, an agent that optimizes expected reward is better positioned to approximate the latent reward structure of an environment than an agent that strictly maximizes instantaneous reward, as in standard Q-learning.
This distinction becomes particularly relevant in environments that contain continuous gradients of a target compound.
However, complications arise when the environment contains regions devoid of the target signal.
During early training, reward assignments in such regions are highly uncertain, causing a Q-learner to maximize an unreliable objective. In contrast, an Expected SARSA learner \cite{Sutton1998} estimates the expectation over possible actions, thereby approximating the policy itself and effectively reducing loss.
Although regularization can partially mitigate this issue, it does not fully resolve the instability that erroneous reward estimates introduce.

In much of contemporary reinforcement learning research — especially in domains with directly observable environments — Q-learning and its variants represent a practical compromise between computational cost and performance.
Our experimental results, however, indicate that the effectiveness of Q-learning depends strongly on the agent's ability to observe rewards that faithfully reflect ground truth.
When reward signals are reliable, a greedy Q-learner is likely to achieve superior performance.
For this reason, we posit that Expected SARSA constitutes a more robust choice in general, with its update rule given by

\[
    Q^*(s,a) = Q(s, a) + \alpha \left [R(s, a, s') +  \frac{\gamma}{n}\sum_{i=1}^{n} Q(s'_i, a_i') - Q(s, a) \right ]
\]

Expected SARSA further allows explicit control over the degree of greediness in the policy.
As $\epsilon$ converges to 1, Sutton and Barto \cite{Sutton1998} demonstrate that the method converges to selecting the action with maximal cumulative reward, thereby becoming equivalent to Q-learning.
This flexibility enables Expected SARSA to combine the benefits of both paradigms by adjusting policy learning according to the reliability of reward estimates and the self-paced learning parameter $\lambda$.
Accordingly, the early stages of training resemble SARSA,
\[
Q^*(s,a) = Q(s, a) + \alpha [R(s, a, s') + \gamma  Q(s', a') - Q(s, a)]\]

\noindent which dampens excessive loss, and gradually transition toward Q-learning,
moderating the high loss as the policy improves:
\[
Q^*(s,a) = Q(s, a) + \alpha \left [R(s, a, s') + \gamma \   \underset{a'}{\max}\ Q(s', a') - Q(s, a)\right ]
\]
This formulation allows the algorithm to operate in both on-policy and off-policy regimes.

A closer inspection reveals that the principal distinction between Expected SARSA and Q-learning lies in replacing reward maximization with reward averaging.
Consequently, Expected SARSA incurs greater computational cost, since the algorithm must evaluate an expectation over actions at each update.
Nevertheless, this added expense yields increased robustness and accuracy during co-training by reducing the volatility that early reward mis-estimation causes.
This advantage is particularly pronounced in the initial phases of learning, when unrewarded trajectories may receive highly inaccurate values due to frequent blank observations.
Because the objective is to infer rewards for unobservable states with greater fidelity, the additional computational burden may be warranted.
Expected SARSA thus provides a principled mechanism for interpolating between on-policy and off-policy learning while offering an inherent means of controlling reward loss.
In our experiments, $\epsilon$ in Expected SARSA increases as a function of the self-paced learning parameter.

In light of the above, simulation showed that Expected SARSA($\lambda$) was more robust to perturbations in the environment, reasoning about blanks, and training parameter changes, so we selected it as the finalist to deploy onto the UAV for evaluation.
We trained both algorithms with the following configuration.
Details on its architecture appear below:

\begin{table}[htb]
    \caption{Q($\lambda$) \& Expected SARSA($\lambda$) Configuration}
    \begin{center}
    \begin{tabular}{|c|c|}
    \hline
    \textbf{Parameter}& \textbf{Value}\\
    \hline
    number of states& 9\\
    number of actions& 7\\
    episodes& 10000\\
    $\gamma$ (discount factor)& 0.9\\
    $\alpha$ (learning rate)& 1e-4\\
    $\epsilon$ (greediness)& 1.0\\
    $\epsilon$-decay& 0.999\\
    $\lambda$ (TD step size)& 0.8\\
    \hline
    \end{tabular}
    \label{tab:sm_rl_params_table}
    \end{center}
\end{table}

\begin{figure}[htb]
  \centering
  \includegraphics[width=170mm]{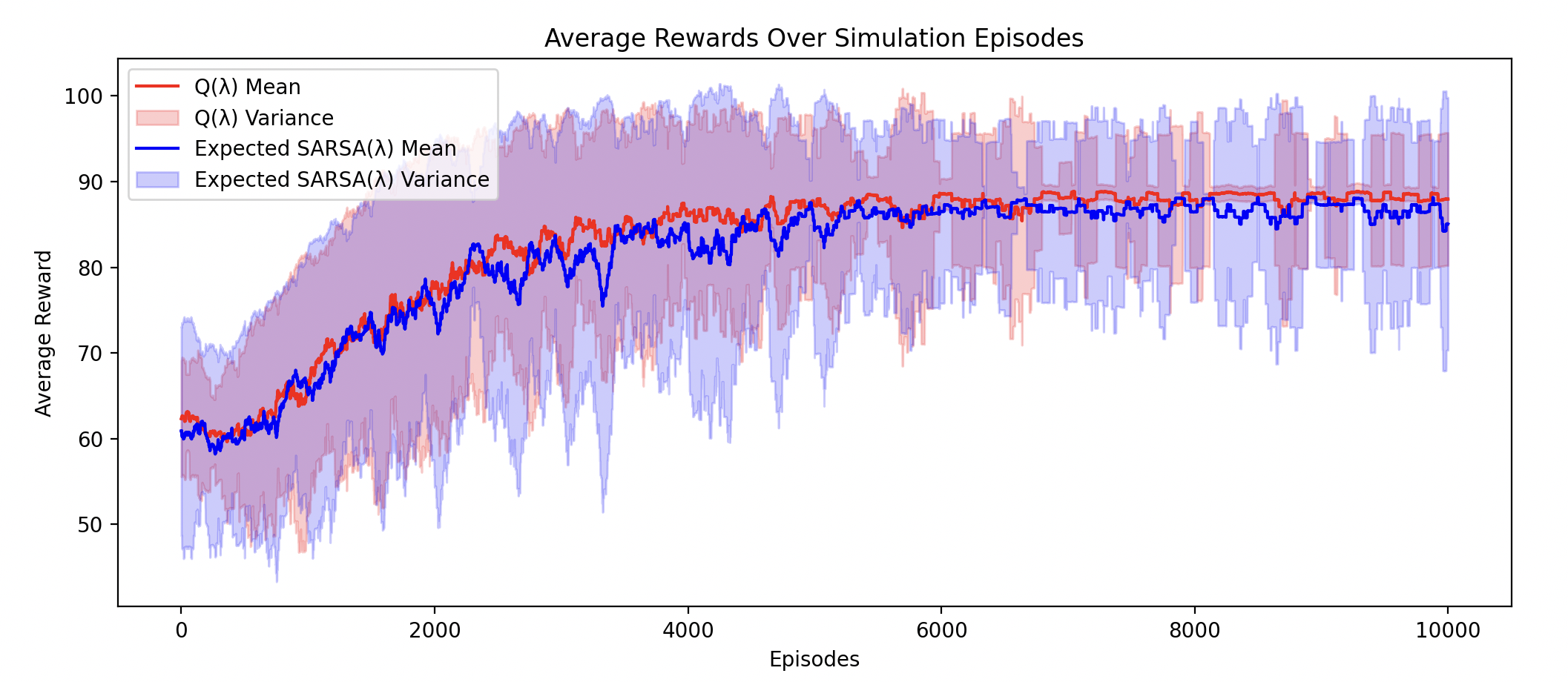}
  \caption{Training losses for both Q($\lambda$) and Expected SARSA($\lambda$) algorithms with no blanks modeled.}
  \label{fig:sm_rl_algos}
\end{figure}

\begin{figure}[htb]
  \centering
  \includegraphics[width=170mm]{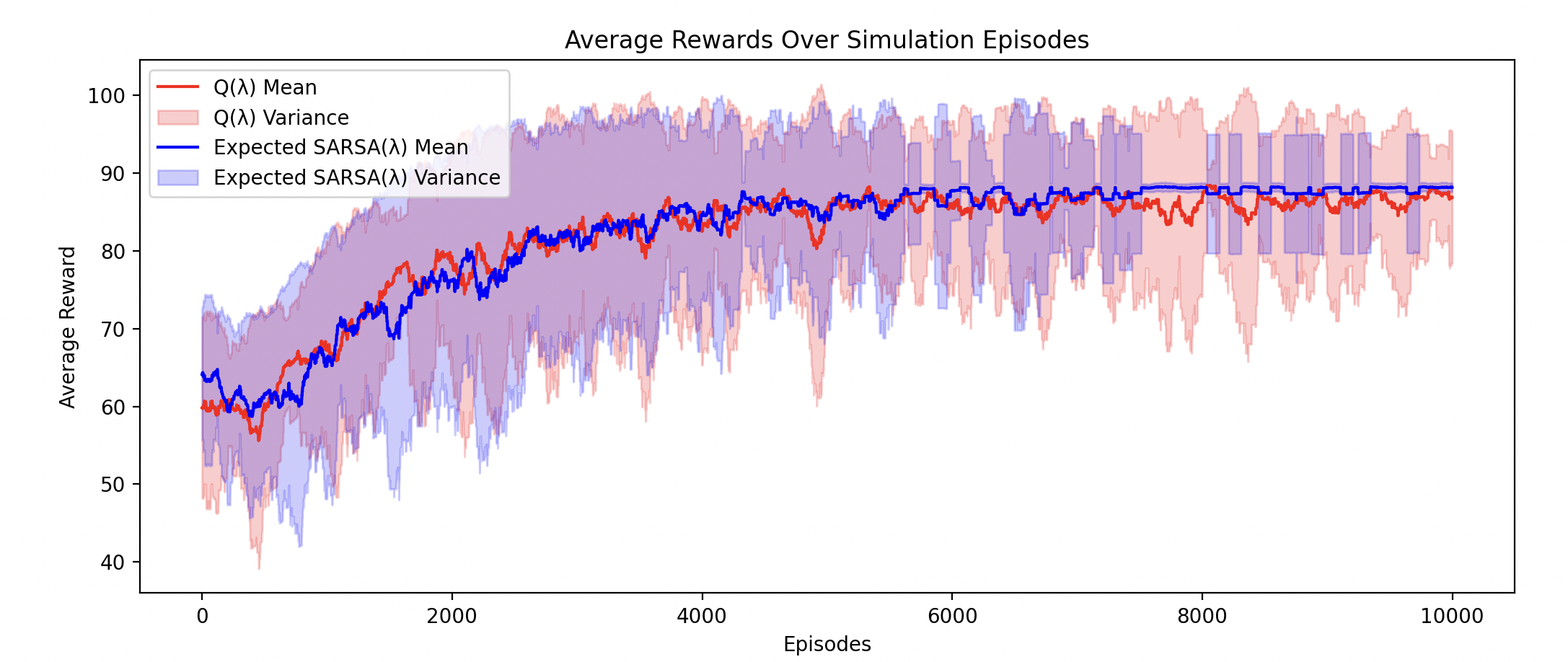}
  \caption{Training losses for both Q($\lambda$) and Expected SARSA($\lambda$) algorithms with blanks modeled.}
  \label{fig:sm_rl_algos_blanks}
\end{figure}

Figures \ref{fig:sm_rl_algos} and \ref{fig:sm_rl_algos_blanks} present the training losses for the final configurations of both algorithms.
Figure \ref{fig:sm_rl_algos} shows how $Q(\lambda)$ performs better in a simulation when no blanks are modeled. 
In other words, no spots within the navigation environment are absent of the target compound.
In reality, this is not probable, and it is prudent to model "blanks" of pockets of air that contain virtually no trace of the target compound \cite{dennler_2025_neuromorophicandolfaction}.
Note the tighter variance bands around $Q(\lambda)$ toward the end of training for the simulation we modeled without blanks.

When we model blanks in the environment, Expected SARSA$(\lambda)$ out-performs $Q(\lambda)$.
We expect that this is due to Expected SARSA being more robust to noise than greedy Q-learning as noted by \cite{France2023, France2024, sutton2018reinforcement}.

\FloatBarrier
At the conclusion of training, a $Q$-value table emerges showing credit assignment to each action. 
Figure \ref{fig:sm_rl_qvalues} below shows the final $Q$-values for the final policy.

\begin{figure}[htb]
  \centering
  \includegraphics[width=140mm]{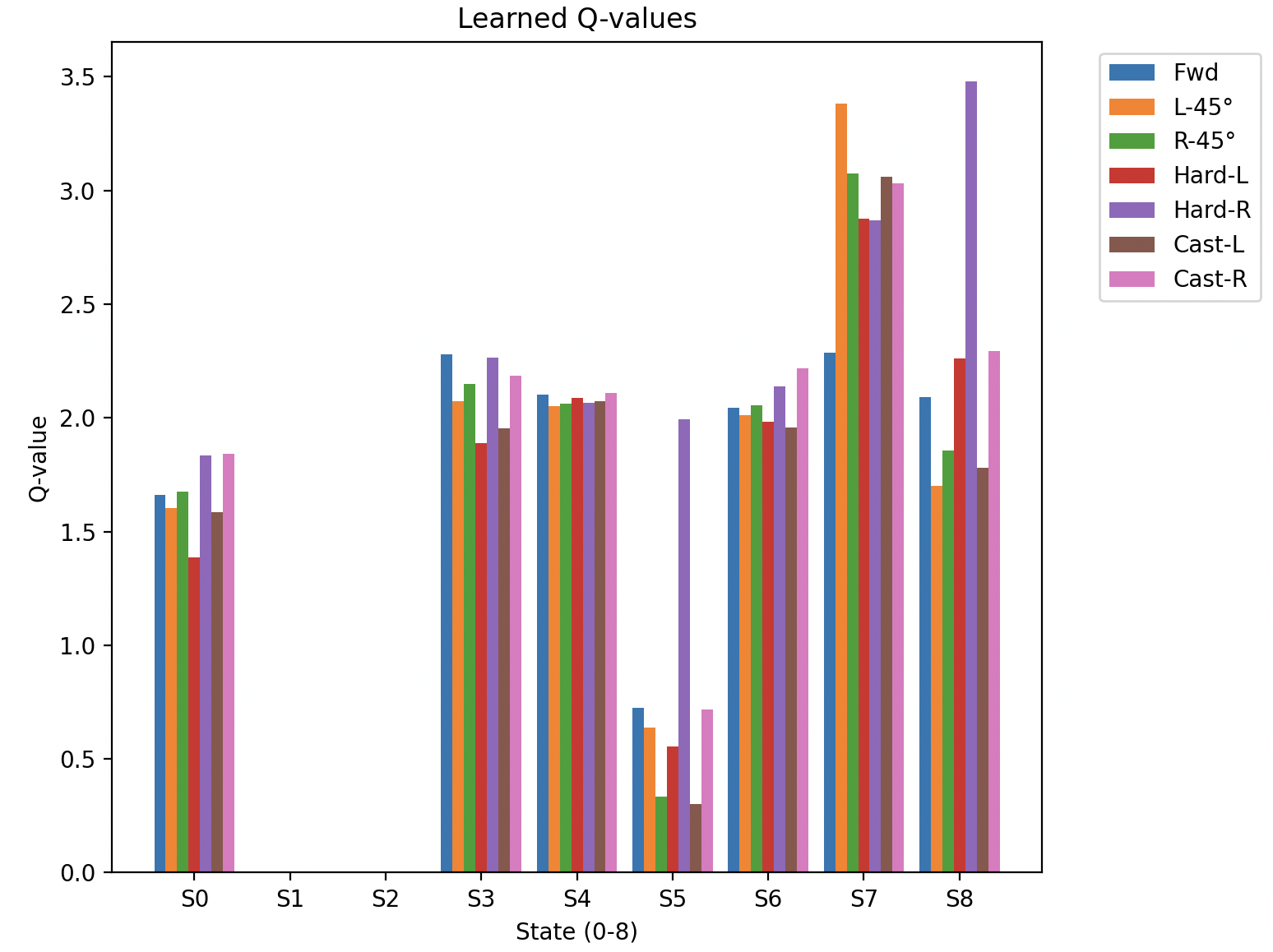}
  \caption{$Q$-values of the final reinforcement learning policy.}
  \label{fig:sm_rl_qvalues}
\end{figure}

\FloatBarrier
\subsection{Plume Model Environment}
\label{sec:sm_plume_env}

We built a plume environment to understand behavioral dynamics of gases over time.
We used the environment as a model to evaluate the effects on air volatility from changing different parameters.
We later used it to help model navigation algorithms for the UAV.
We built the environment on top of the \textit{Gymnasium} framework and included it within the code repository associated with this paper.

We note the configuration to reproduce our results below.

\begin{table*}[htb]
    \caption{Plume Environment Parameters}
    \begin{center}
    \begin{tabular}{|c|c|c|c|}
    \hline
    \textbf{Parameter}& \textbf{Value}& \textbf{Units}& \textbf{Description}\\
    \hline
    Diffusion& 1.0& Dimensionless& Rate of gas dispersion per timestep; conditional on compound\\
    Sparsity& 0.0& Dimensionless& Useful for modeling blanks over long distances; conditional on compound\\
    Temperature& 20.0& Degrees Celsius& Assume constant air temperature due to short durations and distances\\
    Relative Humidity& 50.0& Percentage& Assume constant relative humidity due to short durations and distances\\
    Air Density& 1.225& $kg/m^3$& Assume constant air density due to short durations and distances\\
    Wind Speed& 1.0& $m/s$& Average windspeed along $x$-direction\\
    Emission Rate& 1.0& $kg/s$& Rate of gas emission from source; conditional on compound\\
    Pasquill Stability& D& Class A-G& General air volatility characteristic; conditional on compound\\
    Source Height& 1.0& $m$& We modeled a maximum of 3.0 meters due to flight envelope of UAV\\
    Obstacles& 5& Integer count& We model only square block obstacles\\
    \hline
    \end{tabular}
    \label{tab:paramsTable}
    \end{center}
\end{table*}

\FloatBarrier
\subsection{Olfaction Sensor Breakout Board}
\label{sec:sm_uavbrakout}

\begin{figure}[htb]
  \centering
  \includegraphics[width=100mm]{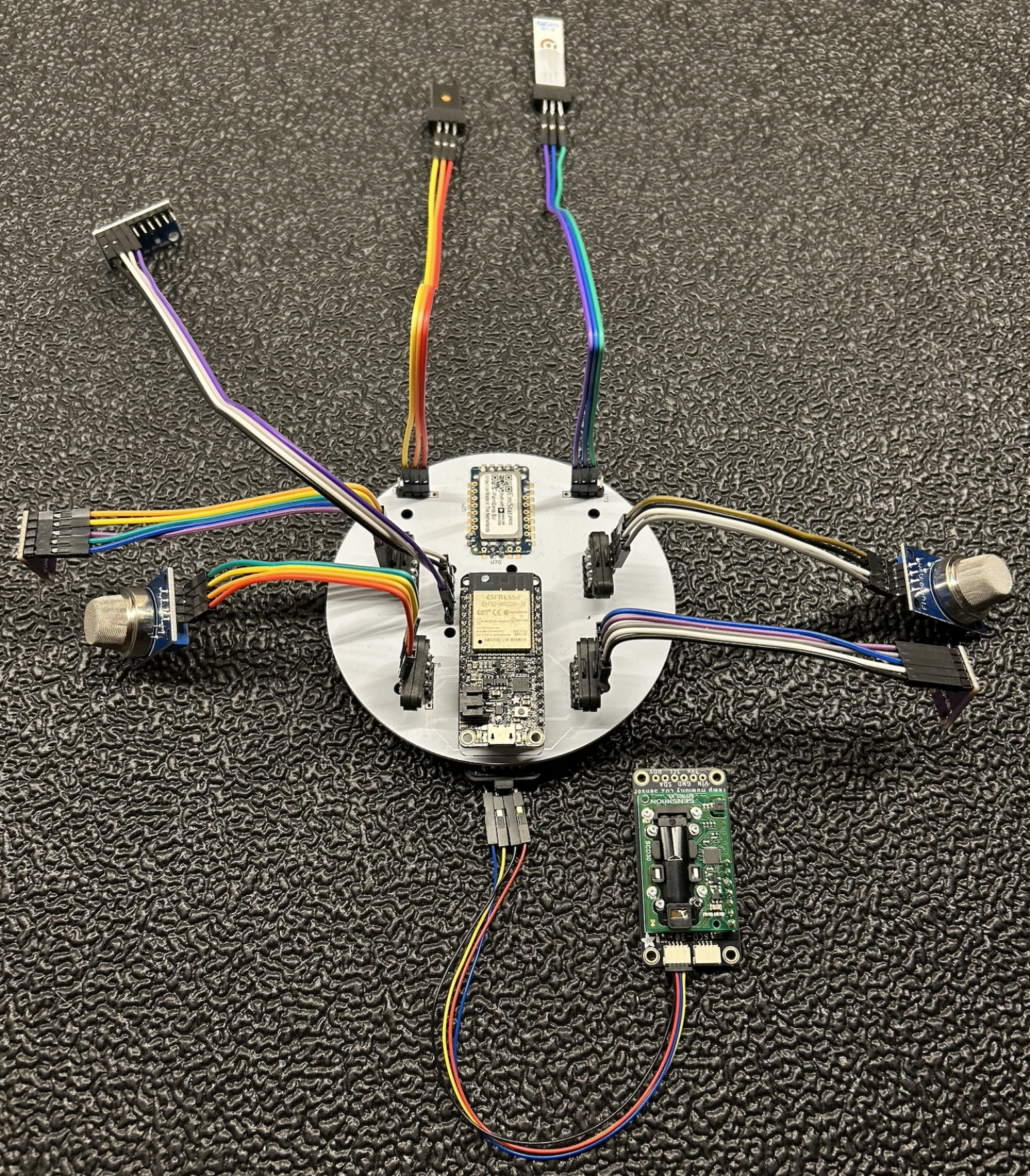}
  \caption{An image of the breakout board we designed to down-select the olfaction sensors for ethanol detection.}
  \label{fig:sm_uavbreakout}
\end{figure}

\begin{figure}[htb]
  \centering
  \includegraphics[width=170mm]{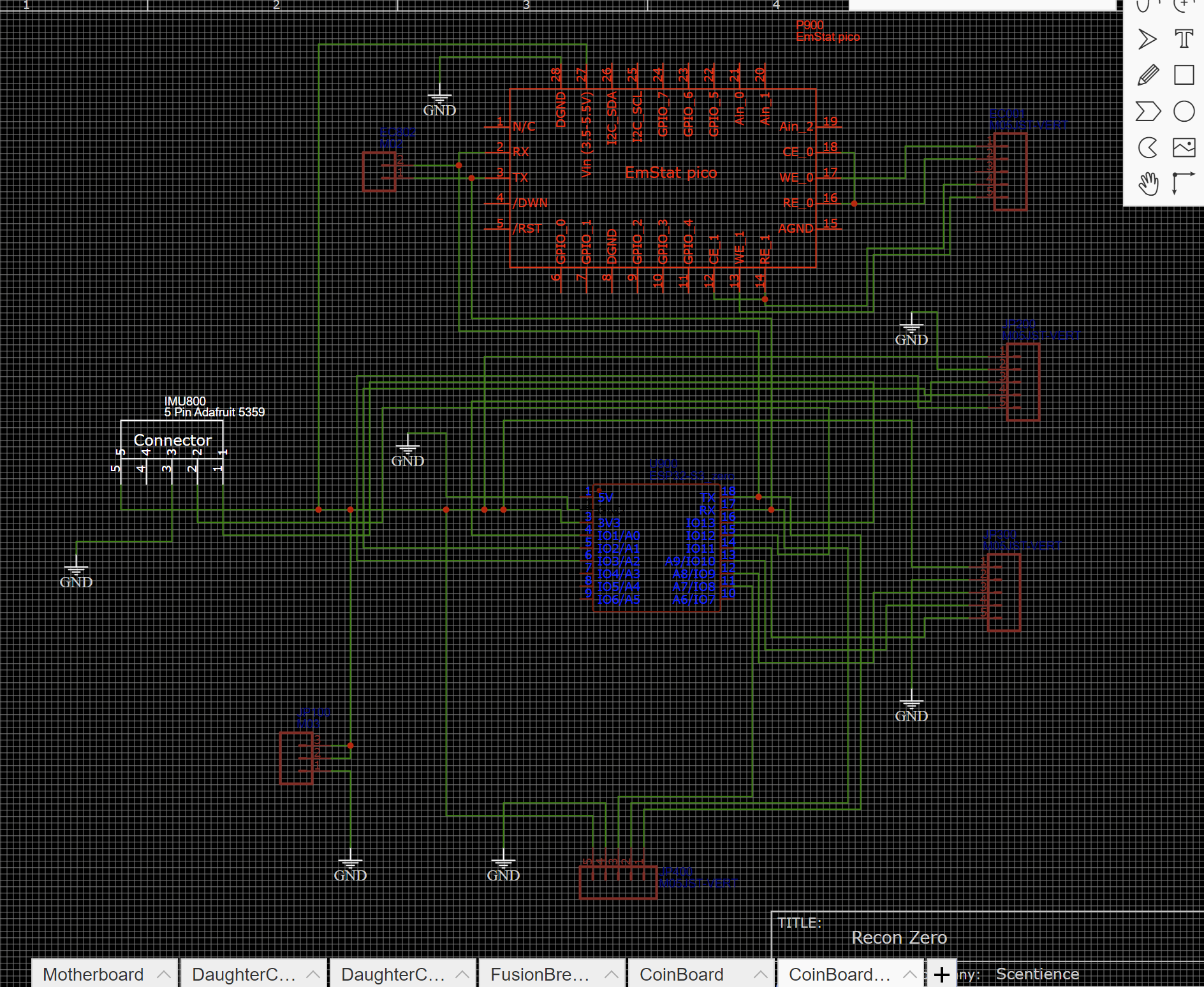}
  \caption{Electrical schematic of the breakout board above. The code repository associated with this manuscript provides Gerber files and full schematics for this board.}
  \label{fig:sm_uavbreakoutschematic}
\end{figure}

Several different olfaction sensors exist.
In order for us to understand which ones were best suited for the detection of ethanol, we designed a breakout board that assessed several olfaction sensors in parallel.
We documented and compared responses to concentration changes in ethanol, leading to our ultimate selection of the metal oxide and electrochemical sensors as the best fit for tracking ethanol.
The code repository associated with this paper contains the electrical schematics and Gerber files for this circuit board in order to encourage olfactory research.
Figure \ref{fig:sm_uavbreakout} shows the final assembled breakout board.
Figure \ref{fig:sm_uavbreakoutschematic} shows the electrical schematic of the board.
We accumulated data for assessing maximum sensor response with this board via the \textit{Scentience} olfactory interface application \cite{scentience2025_olfactory_interface}.

\FloatBarrier
\subsection{Mechanical Modifications to the UAV}
\label{sec:sm_uavcad}
To adapt the UAV for olfactory navigation, we designed a mechanical harness that equipped the machine with the OPU, olfaction sensors, battery, ballast, and infrared sensors.
We designed all components in SolidWorks and iterated several times; the final iterations reside in the code repository under the \textit{/cad} subdirectory.
We followed generally-accepted mechanical design principles for aeronautics.
One can 3D-print the full kit on a single 500mm x 500mm print bed using PLA plastic.
We used a 35\% infill to balance durability and weight.
The code repository contains a full bill of materials.

\begin{figure}[htb]
  \centering
  \includegraphics[width=130mm]{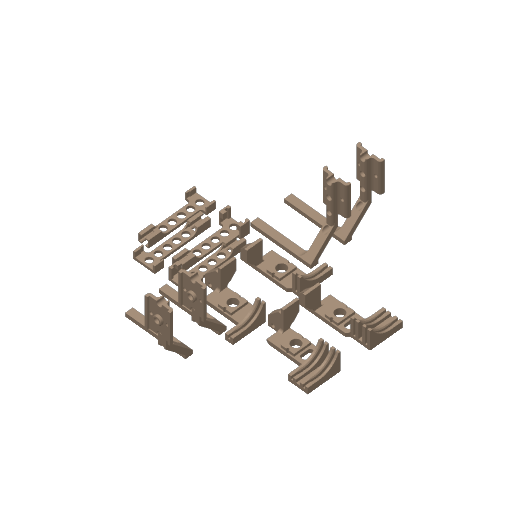}
  \caption{One can additively manufacture the full UAV modification kit for both electrochemical and metal oxide sensing configurations on a single 200mm x 200mm print bed.}
  \label{fig:sm_uavcad}
\end{figure}

\end{document}